%% file: main.tex
\documentclass{article}

\usepackage[final]{neurips_data_2024}

\usepackage{makecell}
\usepackage[
  separate-uncertainty = true,
  multi-part-units = repeat
]{siunitx}
\usepackage[utf8]{inputenc} %
\usepackage[T1]{fontenc}    %
\usepackage{hyperref}       %
\usepackage{url}            %
\usepackage{booktabs}       %
\usepackage{amsfonts}       %
\usepackage{nicefrac}       %
\usepackage{microtype}      %
\usepackage{xcolor}         %
\usepackage{longtable}
\usepackage{graphicx}
\usepackage{subcaption}

\usepackage[english]{babel}

\usepackage{amsmath}
\usepackage{tabularx} 
\usepackage{listings}

\title{A Careful Examination of Large Language Model Performance on Grade School Arithmetic}

\author{
  Hugh Zhang\thanks{Correspondence to hugh.zhang@scale.com 
  \,\, $^\dagger$equal senior authorship} \And
  Jeff Da \And
  Dean Lee \And 
  Vaughn Robinson \And
  Catherine Wu \And
  Will Song \And
  Tiffany Zhao \And
  Pranav Raja \And
  Charlotte Zhuang \And
  Dylan Slack \And
  Qin Lyu \And
  Sean Hendryx \And
  Russell Kaplan \And
  Michele (Mike) Lunati$^\dagger$ \And
   Summer Yue$^\dagger$
}
\date{}

\newcommand{\gsm}[0]{GSM1k}
\renewcommand{\cite}[1]{(\citet{#1})}

\begin{document}
\maketitle
\begin{center}
{\LARGE Scale AI} %
\vspace{1cm} %
\end{center}

\begin{abstract}
\input{paper/abstract}
\end{abstract}

\input{paper/intro}
\input{paper/related}
\input{paper/methodology}

\input{paper/results}

\input{paper/analysis}
\input{paper/conclusion}

\bibliographystyle{plainnat}
\bibliography{references}

\input{paper/appendix}

\end{document}

%% file: paper/abstract.tex
Large language models (LLMs) have achieved impressive success on many benchmarks for mathematical reasoning.
However, there is growing concern that some of this performance actually reflects dataset contamination, where data closely resembling benchmark questions leaks into the training data, instead of true reasoning ability.
To investigate this claim rigorously, we commission \emph{Grade School Math 1000} (\gsm). \gsm{} is designed to mirror the style and complexity of the established GSM8k benchmark,
the gold standard for measuring elementary mathematical reasoning. We ensure that the two benchmarks are comparable across important metrics such as human solve rates, number of steps in solution, answer magnitude, and more.
When evaluating leading open- and closed-source LLMs on \gsm, we observe accuracy drops of up to 8\%, with several families of models showing evidence of systematic overfitting across almost all model sizes.
Further analysis suggests a positive relationship (Spearman's $r^2=0.36$) between a model's probability of generating an example from GSM8k and its performance gap between GSM8k and \gsm, suggesting that some models may have partially memorized GSM8k. 
Nevertheless, many models, especially those on the frontier, show minimal signs of overfitting, and all models broadly demonstrate generalization to novel math problems guaranteed to not be in their training data.

%% file: paper/intro.tex
\section{Introduction}
\begin{figure}[ht!]
    \centering
    \includegraphics[width=1\linewidth]{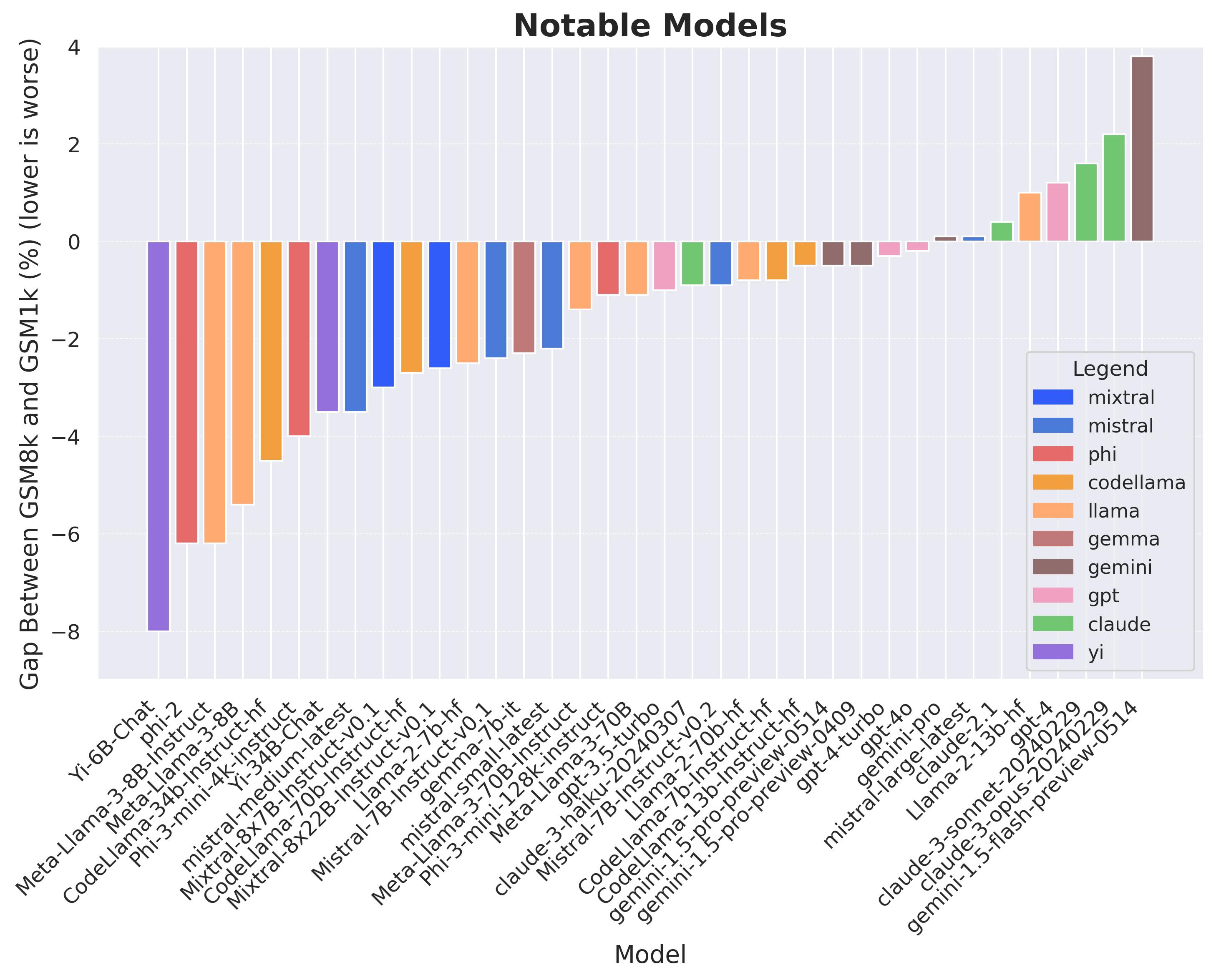}
    \caption{Notable models arranged by their drop in performance between GSM8k and GSM1k (lower is worse). We notice that Phi, Mistral and some models in the Llama family seem to be overfitting GSM8k, while models such as Gemini, GPT, and Claude show little to no signs of overfitting.}
    \label{fig:topoverfit}
\end{figure}

Improving reasoning in large language models (LLMs) is one of the most important directions of current research.
As such, proper benchmarking of current LLM abilities is paramount for ensuring progress continues in the correct direction.
Currently, the field typically relies on public benchmarks such as GSM8k \cite{cobbe2021training}, MATH \cite{hendrycks2021measuring}, MBPP \cite{austin2021program}, HumanEval \cite{chen2021evaluating}, SWEBench \cite{jimenez2024swebench}).
However, because LLMs are trained on large corpora of data scraped from the Internet, there are major concerns that such benchmarks may inadvertently include examples that closely resemble the questions found in such benchmarks.
This contamination may result in models having weaker reasoning capabilities than otherwise believed, due to simply being able to repeat the correct answer that it previously encountered during pre- or post- training.
To properly investigate the reasoning abilities of models, we commission \gsm{}, a newly constructed collection of 1205 grade school level math problems designed to mirror that of GSM8k.
We take extensive efforts to ensure that GSM1k has a similar distribution of difficulty to GSM8k to ensure an apples-to-apples comparison. These efforts are described in Section~\ref{sec:methodology}, alongside a detailed description of the data creation process. 
To mitigate worries about data contamination, we created \gsm{} solely with human annotators, without assistance from any LLM or other synthetic data source. 
\begin{figure}[ht]
\centering
\begin{tabularx}{\textwidth}{lX}
\toprule
\textbf{Dataset} & \textbf{Example} \\
\midrule
GSM8k & James writes a 3-page letter to 2 different friends twice a week. How many pages does he write a year? \\
\addlinespace[1em]  %
\gsm \,\, (ours) & Lee bought 6 shares of Delta stock at \$40 per share. If he wants to make \$24 from this trade, how much should Delta stock be per share when he sells? \\
\bottomrule
\end{tabularx}
\caption{Example from both the GSM8k dataset and the new \gsm{} dataset (ours).
We also provide an additional 50 examples from \gsm{} in Appendix~\ref{app:50examples}.}
\label{fig:example_tasks}
\end{figure}

We benchmark leading open- and closed-source LLMs on \gsm, including GPT-4 \cite{openai2024gpt4}, Gemini \cite{geminiteam2024gemini}, Claude, Mistral \cite{jiang2024mixtral, jiang2023mistral}, Llama \cite{touvron2023llamaa, touvron2023llamab}, Phi \cite{gunasekar2023textbooks, abdin2024phi3} and many more.
Our analysis confirms the widespread suspicion in the field that many models are contaminated by benchmark data, with the worst models performing 8\% worse on GSM1k compared to GSM8k. Additionally, our results suggest that several families of models show consistent evidence of overfitting for nearly all model versions and sizes.
Further analysis finds a positive relationship (Spearman's $r^2=0.36$) between a model's probability of generating examples from GSM8k and its performance gap between GSM8k and GSM1k, strongly suggesting that one important component of this overfitting is that models have partially memorized examples from GSM8k.
Nevertheless, our results find that all frontier models show minimal signs of overfitting.
Additionally, we also find that all models, including the most overfit ones, are still capable of successfully generalizing to new mathematical grade school problems, albeit occasionally at lower rates than their benchmark numbers would suggest.

We do not intend to release GSM1k publicly at this time to prevent a similar problem of data contamination occurring in the future. However, we plan to run recurring evaluations of all major open- and closed- source releases and to continually update our results. We will also open source our entire evaluation code so that the public version of our results can be reproduced. Additionally, we commit to open sourcing the entire benchmark when either 1) the top open source models score over 95\% on \gsm{} or 2) June 2025, whichever comes earlier. See Section~\ref{sec:methodology} for precise release criteria.

%% file: paper/related.tex
\section{Related Work}
A major inspiration of this work was the celebrated study on overfitting done on ImageNet classifiers in 2019 \cite{recht2019imagenet}. This work measured overfitting in ImageNet by creating new versions of CIFAR10 and ImageNet and measuring the performance gap between the public test set and the newly created sets they constructed. In this work, we do a similar analysis on GSM8k, one of the leading benchmarks for mathematical reasoning.
\gsm{} is modelled after the GSM8k dataset \cite{cobbe2021training}, released by OpenAI in 2021, which consists of 8.5k grade school math problems. Each problem is designed to be solvable using only basic arithmetic operations ($+$, $-$, $\times$, $\div$) with a difficulty level appropriate for grade school students.
As of June 2024, top models report benchmark accuracies of over 95\% \cite{geminiteam2024gemini}.
Other popular benchmarks for reasoning include MATH \cite{hendrycks2021measuring} , MMLU \cite{hendrycks2021measuringb}, GPQA \cite{rein2023gpqa}.
\subsection{Data Contamination}
Because data contamination is a well known issue in the field \cite{balloccu2024leak, magar2022data, sainz2023nlp, jacovi2023stop, xu2024benchmarking}, model builders will frequently take great pains to minimize the likelihood of data contamination. 
For example, it is common to remove all data with too high of an n-gram overlap with the benchmark data \cite{brown2020language}. Additionally, methods such as using embedding similarity attempt to remove all contaminated data that is too similar in embedding space to the dataset \cite{shi2024detecting}.

\citet{xu2024benchmarking} propose using similar variants of a benchmark questions to detect if models favor the original wording as a proxy for data contamination.
\citet{srivastava2024functional} propose functional evaluations, where benchmarks are written in the form of functions that can generate an infinite number of specific evaluation datapoints, each with slightly different numbers.
In this setup, whenever a language model is evaluated, functional evaluations generate a specific problem instance to evaluate the model on, which is then never used again.
This reduces the worry of data contamination by ensuring that no datapoint is ever used twice.
Like ours, their results indicate the LLMs may be severely overfit on benchmark data.
The main advantage of our approach over a purely function based evaluation is that functional evaluations can only generate a tiny portion of the full problem space by producing variations of the same problem with slightly different numerical values. Their results also suggest substantial amounts of data contamination, including for frontier models, in the MATH dataset.

%% file: paper/methodology.tex
\section{GSM1k}
\label{sec:methodology}

\gsm{} consists of 1205 problems requiring only elementary mathematical reasoning to solve. 
We created \gsm{} using human annotators. Annotators were prompted with 3 example GSM8k problems and asked to produce novel problems of a similar difficulty level. The precise instructions and UI given to the annotators is available in Appendix~\ref{app:annotator_instructions}.
All problem annotators were instructed to create problems solvable with only basic arithmetic (addition, subtraction, multiplication, and division) and which did not require any advanced math concepts. As is the case with GSM8k, all problem solutions are positive integers\footnote{GSM8k has a few problems, likely errors, for which this is not the case.}.
No language models were used to construct this dataset.

To prevent data contamination concerns with \gsm, we do not intend to release the dataset publicly at this time.
However, we commit to releasing the full \gsm{} dataset when at least one of the two following conditions have passed, whichever comes earlier. 1) Three open-source models with different pre-trained foundational model lineages reach 95\% accuracy on GSM1k. 2) June 2025.
At such a point, we believe that grade school mathematics will likely no longer be difficult enough to materially benchmark model releases and commit to releasing all data into the public domain under the MIT license.
Additionally, to evaluate proprietary models, we were required to send over the dataset via API. Our belief is that model providers typically do not use such datapoints for model training.
Nevertheless, in case GSM1k data is leaked through such means, we also hold out a small number of data points that have passed all quality checks but do not appear in the final \gsm{} dataset.
This data will also be released alongside \gsm{} upon final release.
We encourage future benchmarks to follow a similar pattern, where they are not released publicly lest they be gamed, but are precommitted to be released at a future date or upon a future condition.
As part of this release, we will also open source our evaluation framework, which is based off of a fork of the LM Evaluation Harness by EleutherAI \cite{eval-harness}. 

Finally, while we undertook extensive efforts to ensure maximum similarity between GSM8k and GSM1k, these results are only an approximation of an ideal world in which the test set of GSM8k was instead not publicly released and used for evaluations. We would recommend reading all results with the understanding that GSM8k and GSM1k are only highly similar, but not identically distributed despite all our efforts below.
\subsection{Quality Checks}
All questions passed through a total of 3 review layers. After initial creation, each task was manually reviewed by a subset of trusted annotators selected for strong past performance.
These reviewers checked both for correctness as well as ensuring problems contained only grade school level math and proper formatting.
To ensure that questions were answered correctly, we also do a second review layer by having an independent set of data annotators solve each question \emph{without seeing the intended solution}.
If this second solve produced a different answer to that of the initial solve, we discarded the problem.
Finally, all problems were reviewed by a special team within Scale responsible for conducting general quality audits for data production. Out of a total of $2108$ initial problems, $1419$ passed the second solve stage and $1375$ passed the general quality audit.
\subsection{Matching the Difficulty Distribution of GSM8k}
\label{sec:difficulty}
One important axis of recreating a benchmark is ensuring that new problems have a comparable difficulty to the original benchmark. To construct problems of difficulty $N$, we requested annotators to construct problems with $N$ required resolution steps and prompted them with 3 examples from GSM8k with estimated difficulty $N$. The distribution of problems requested from annotators matched the estimated distribution in GSM8k.
Difficulty is tricky to measure precisely, so we used an estimate based on the number of operations needed to solve the problem. This was extracted programmatically by counting the number of ``calculator'' tags in the problem solution. However, as not all problem solutions were formatted consistently, this estimate is only a rough estimate of actual difficulty. Additionally, the number of resolution steps in a problem does not necessarily directly correlate with the true level of problem difficulty.

\begin{figure}[ht!]
    \centering
    \includegraphics[width=0.8\textwidth]{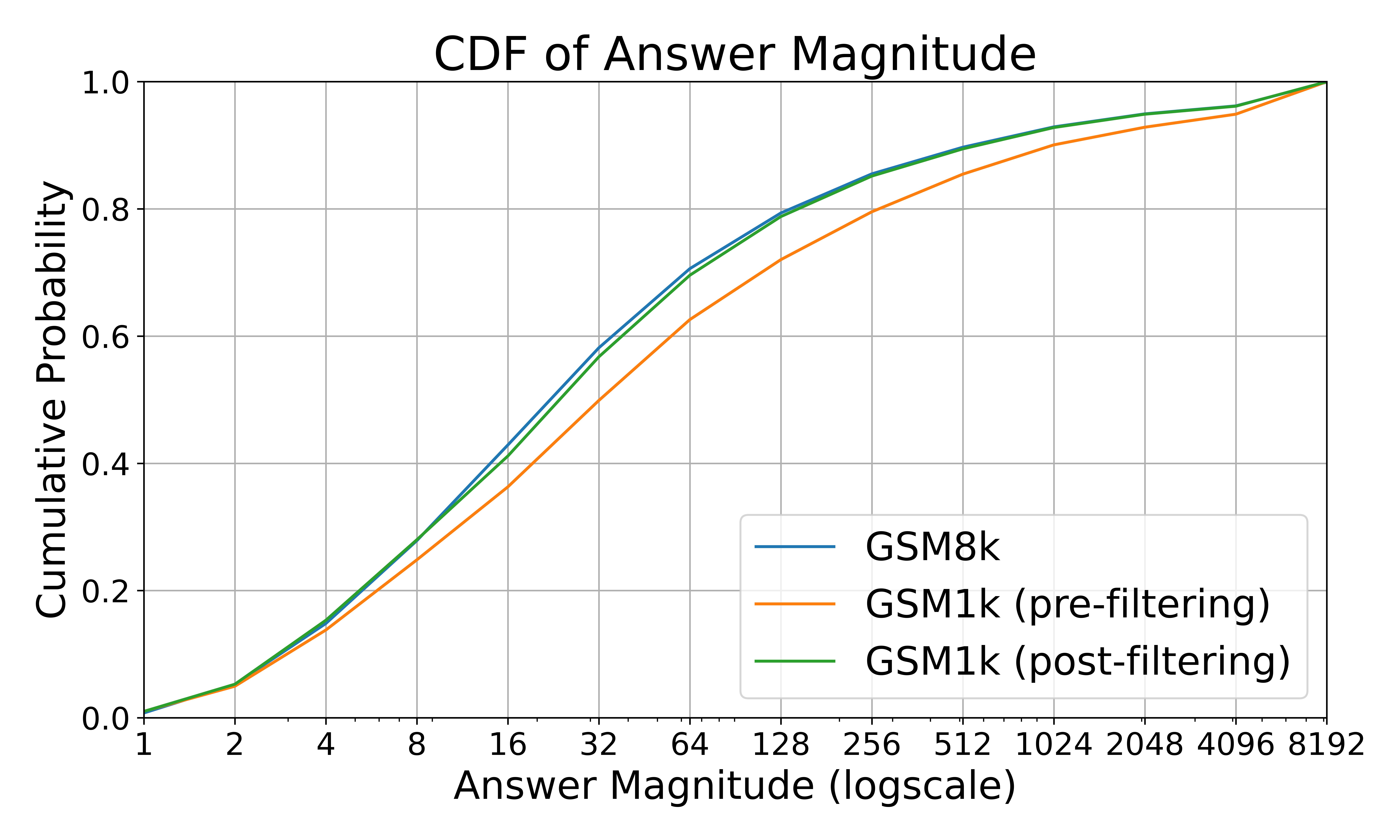}
    \caption{As the final step, we select 1205 problems to match the answer magnitude distribution of GSM8k as much as possible. The remaining problems are discarded and not included in the final dataset. Before discarding, we find that our generated problems tend to have slightly larger answers.}
    \label{fig:cdf}
\end{figure}

Past work has also found that LLMs struggle with problems with larger numbers \cite{gao2023pal} even if they can solve otherwise identical problems with smaller numbers. To remove this as a potential confounding variable, our final processing step is to discard candidate problems from GSM1k so that the answer magnitude distributions of GSM8k and GSM1k are as similar as possible. This selection process is described in Figure~\ref{fig:cdf}. GSM1k consists of the 1205 problems that survive this final winnowing.
Additionally, we run several checks to ensure that our efforts to match benchmark difficulty were successful.
\subsubsection{Human Differentiation Rates}
The first test we run is human distinguishability. We present human annotators with a set of five questions, four of which were randomly selected from the original GSM8k dataset and one of which was selected from the newly created \gsm{} dataset, and rewarded annotators for finding the odd one out.
In an audit conducted using 19 annotators who were not involved in the problem creation process, we found that annotators were able to correctly identify the lone \gsm{} example 21.83\% of the time out of 1205 attempts (20\% is pure chance). Separately, we also tested several paper authors who had not yet seen the data and they were also unable to perform much better than random. This suggests minimal differences between GSM8k and GSM1k, at least as measured by the human eye.
\subsubsection{Human Solve Rates}
To ensure similar solve rates, we also asked annotators to solve questions under time pressure. 
14 annotators who had not participated in the problem creation process attempted to solve as many GSM8k problems as they could in 15 minutes and were rewarded based on the number of problems they solved. We repeated this exact setup for GSM1k. Annotators were able to solve an average of $4.07 \pm 0.93$ problems on the GSM8k dataset. They were able to solve $4.36 \pm 1.11$ problems on the GSM1k dataset, where the error rates are the standard deviations of the evaluation. This suggests that GSM1k is comparable in difficulty (and perhaps even slightly easier) than GSM8k. As such, substantial decreases in model accuracy on GSM1k compared to GSM8k are likely not explainable due to differences in dataset difficulty.
\subsubsection{LLM Solve Rates}
Finally, we sanity check our results by measuring solve rates of several models that are known to not be contaminated by GSM8k due to being released before the publication of the GSM8k dataset. Due to the relative scarcity of LLMs trained only on pre-2021 data, we evaluate only GPT-NeoX-20B \cite{black2022gptneox20b} and GPT-2 \cite{radford2019language}. For these two language models, we find minimal difference between their solve rates of GSM8k and GSM1k (Figure~\ref{fig:badaccuracyplot}).

%% file: paper/results.tex
\section{Results}
\label{sec:results}
\begin{figure}[h!]
  \includegraphics[width=\linewidth]{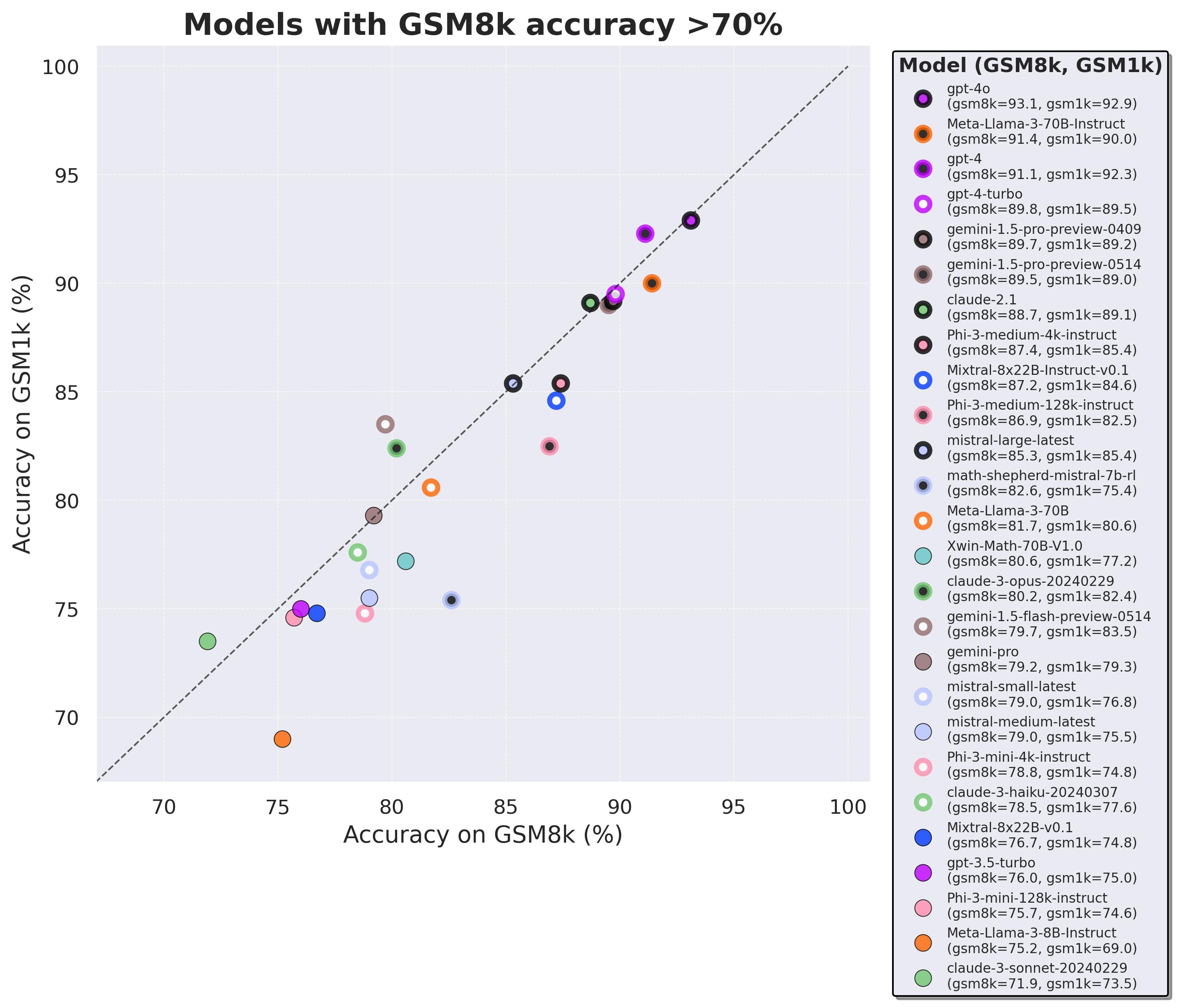}
  \caption{Models with over 70\% accuracy on GSM8k compared to the line of no overfit. This plot is zoomed into the relevant sections (70-100\% accuracy). Note that some models, especially the Claude family, perform above the 45 degree line, which is consistent with our findings in Section~\ref{sec:methodology} that GSM1k is slightly easier than GSM8k. In contrast, many other models lie well below this line.} 
  \label{fig:goodaccuracyplot}
\end{figure}
To evaluate models, we use a fork of EleutherAI's LM Evaluation Harness with minor modifications.
We use the default settings for evaluation, except for increasing the maximum number of allowed generated tokens from 256 to 1000, as we notice that the default setting did not allow some models to complete their full chain-of-thought reasoning before being truncated. Both GSM8k and GSM1k questions are run with the same prompt of using 5 randomly drawn examples from the GSM8k train set, as is standard in the field. An example prompt is provided in Appendix~\ref{app:prompt}. All open-source models are evaluated at temperature 0 for reproducibility. For open source models, we use vLLM to speed up model inference if a model is compatible with the library. Otherwise, we default to inference using standard HuggingFace libraries. Closed-source models were queried through the LiteLLM library which unifies the API call format for all proprietary models evaluated. All API model results were from queries between April 16 - July 10, 2024 and use the default settings.

LM Evaluation Harness uses an automatic evaluation method which extracts the last numeric answer in the response and compares this to the correct answer. However, in some cases, models will produce ``correct'' answers in a format that do not match the given examples, resulting in their answers being marked as incorrect. To explore the effect of this on the results, we run an ablation where we select a subset of models and use human annotation to manually extract answers that are not correctly formatted (Appendix~\ref{app:manual}). We do not find major changes in our findings for the models examined.

As model benchmark performance is highly dependent on choice of prompt and evaluation setting, our reported GSM8k numbers may occasionally be below the reported model benchmark numbers, as we use a standardized setting for all models instead of the prompt that maximizes each individual model's performance. Additionally, we explore the effect of different prompt formulations with several ablations. In Appendix~\ref{app:alternative}, we report results with an alternative prompting format that uses non-GSM8k examples as n-shot examples and a slightly different answer phrasing. Additionally, we explore the effect of varying the number and source of the n-shot examples used in Appendix~\ref{app:ablation_alt} and \ref{app:ablation_n}.
While the precise benchmark accuracies vary depending on the setup, we find that the general trends of overfitting hold consistently across our ablations.
We will release the full evaluation code for transparency.

In addition to evaluating widely known models, we additionally evaluate several lesser known models that sit near the top of the OpenLLMLeaderboard and discover evidence of Goodhart's law: many of these models perform substantially worse on GSM1k, suggesting that they are primarily gaming the GSM8k benchmark rather than improving model reasoning capabilities. The full set of results, including the performance table for all models, can be found in Appendix~\ref{app:table}. For fair comparison, we partition the models by performance on GSM8k and compare them to other models which perform similarly (Figures~\ref{fig:goodaccuracyplot}, \ref{fig:medaccuracyplot}, \ref{fig:badaccuracyplot}).

%% file: paper/analysis.tex
\section{Analysis}
\begin{figure}[th!]
    \centering
    \includegraphics[width=1\linewidth]{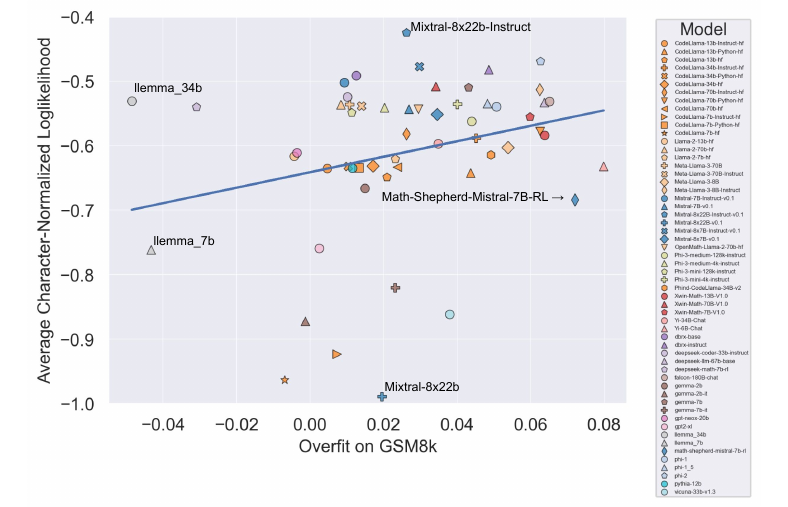}
    \caption{Comparison between overfit on GSM8k (x-axis) and average sequence-level log-likelihood on the GSM8k test set (y-axis). We find that there is a correlation between overfit on GSM8k and sequence-level log-likelihood, suggesting that, in general, models that have a high overfit generally have a higher probability of generating the test set. This suggests that some of the GSM8k test set may have leaked into the model training data. The line of best fit is in blue. Additionally, we highlight 5 ``outlier'' models which we discuss further with Lesson 4.}
    \label{fig:llgraph}
\end{figure}
The interpretation of evaluation results, like the interpretations of dreams, is often a very subjective endeavor. While we report our objective results in Section~\ref{sec:results} and Appendix~\ref{app:table}, here we describe four major takeaways from interpreting the results in a more subjective manner. 
\subsection{Lesson 1: Some Model Families are Systematically Overfit}
While it is difficult to draw conclusions from singular data points or model releases, examining a family of models and observing a pattern of overfitting enables us to make more definitive statements.
Several families of models, including the Phi and Mistral families of models, show systematic tendencies to perform stronger on GSM8k compared to GSM1k for almost every release and scale of models.
Other model families, such as Yi, Xwin, Gemma and CodeLlama also show this pattern to a lesser extent.

\subsection{Lesson 2: Other Models, Especially Frontier Models, Show No Signs of Overfitting}
Nevertheless, we find that many models, through all regions of performance, show minimal signs of being overfit. In particular, we find that all frontier or close-to-frontier models (including the proprietary Mistral Large) appear to perform similarly on both GSM8k and GSM1k. We posit two potential hypotheses for this: 1) frontier models have sufficiently advanced reasoning capability so that they can generalize to new problems even if they have already seen GSM8k problems in their training set, 2) frontier model builders may be more careful about data contamination.

While it is impossible to know for certain without looking at the training set for each model, one piece of evidence in favor of the former is that Mistral Large is the \emph{only} model in the Mistral family to show no signs of overfitting. Since the hypothesis that Mistral took unique care in ensuring that only their largest model was free from data contamination seems unlikely, we lean instead towards the hypothesis that sufficiently strong LLMs also learn elementary reasoning ability during training. If a model learns strong enough reasoning capabilities to solve problems of a given difficulty, it will be able to generalize to new problems even if GSM8k has appeared in its training set.

\subsection{Lesson 3: Overfit Models Are Still Capable of Reasoning}
One worry about model overfitting is that models are incapable of reasoning and are only memorizing answers seen in the training data. Our results do not support this conjecture. The fact that a model is overfit does not mean that it is poor at reasoning, merely that it is not as good as the benchmarks might indicate it to be. In fact, we find that many of the most overfit models are still capable of reasoning and solving novel problems. For example, while Phi-2 has a 6\% drop in accuracy between GSM8k and GSM1k, we find that it is still able to correctly solve over half of GSM1k problems -- which are certain to not have appeared in its training distribution. This performance is similar to that of much larger models such as Llama2-70B, which contains over 25x as many parameters. Similarly, Mistral models remain some of the strongest open source models, even accounting for their overfitting. This provides additional evidence for our lesson that sufficiently strong models learn elementary reasoning, even if benchmark data accidentally leaked into the training distribution, as is likely to be the case for the most overfit models.
\subsection{Lesson 4: Data Contamination Is Likely Not The Full Explanation for Overfitting}
A priori, a natural hypothesis is that the primary cause for overfitting is data contamination, e.g. that the test set was leaked in the pre-training or instruction fine-tuning part of the model creation.
Previous work has suggested that models put higher log-likelihoods on data that they have seen during training \cite{carlini2023quantifying}. We test the hypothesis that data contamination is the cause of overfitting by measuring a model's probability of generating an example from the GSM8k test set and comparing it to how overfit it is on GSM8k vs GSM1k, using the assumption that a model's probability of generating the GSM8k test set is a proxy for whether the sequence is likely to have appeared in the training set. We normalize by $c$, the number of characters in the sequence, to make the log-likelihood calculations comparable between sequences and models with different tokenizers. Formally, we have:
\begin{equation}
\frac{1}{c} \sum_{i} \log p(x_i | x_{< i})
\label{eq:loglikelihood}
\end{equation}
with $c$ being the number of characters in the sequence.
Figure~\ref{fig:llgraph} shows the result of this plot against the gap between GSM8k and GSM1k performance.
We indeed find a positive relationship between the two values. We observe a Spearman's rank correlation of 0.36 between the per-character log-likelihood of generating GSM8k and the performance gap between GSM8k and GSM1k $(p = 0.03)$, and the relationship suggests that every percentage point difference in GSM8k and GSM1k performance is associated with an increase of $1.2 \times 10^{-2}$ in the per-character log-likelihood. This result suggests that some of the reason for overfitting is due to partial memorization of the test set. For completeness, we also report the standard Pearson $r^2=0.26$ and the Kendall's $\tau$ of 0.29, but note that Pearson $r^2$ is not the ideal metric due to the curve-of-best-fit not appearing linear.

Nevertheless, data contamination is likely not the full story. We observe this via the presence of several outliers, which cause the $r^2=0.36$ value to be relatively low. Examining these outliers carefully reveals that the model with the lowest per-character log-likelihood (Mixtral-8x22b) and the model with the highest per-character log-likelihood (Mixtral-8x22b-Instruct) are not only variations of the same model, but also have similar levels of overfit \cite{jiang2024mixtral}. Perhaps more intriguingly, one the most overfit models we discovered (Math-Shepherd-Mistral-7B-RL \cite{yu2023metamath}) had a relatively low per-character log-likelihood. Math Shepherd trains a reward model on process level data using synthetic data. As such, we hypothesize that the reward modelling process may have leaked information about the correct reasoning chains for GSM8k even if the problems themselves did not ever appear in the dataset. Finally, we observe that the Llema models \cite{azerbayev2024llemma} have both high log-likelihoods and minimal overfit. These models are open-sourced alongside their training data, and the authors report finding a very small number of GSM8k examples in the training corpus. Nevertheless, they also find (and our study supports) that these few instances do not lead to overfitting. The existence of these outliers suggests that overfitting on GSM8k is not purely due to data contamination, but rather may be through other indirect means, such as model builders collecting data similar in nature to benchmarks as training data or selecting final model checkpoints based on performance on benchmarks, even if the model itself may have not seen the GSM8k dataset at any point via training. Conversely, the reverse is also true: small amounts of data contamination do not necessarily lead to overfitting.

%% file: paper/conclusion.tex
\section{Discussion}
We create \gsm, a novel dataset designed to measure LLM overfitting on GSM8k. When benchmarking leading open- and closed-source models, we find substantial evidence that many models have been contaminated by benchmark data, with models showing performance drops of up to 8\% accuracy. Additionally, we find that several model families show consistent overfitting across almost all model sizes and versions. An extended analysis reveals a positive relationship between a model's likelihood of generating data points in GSM8k and its performance difference between GSM8k and GSM1k, suggesting evidence of data contamination as one of the underlying causes. Nevertheless, we find that frontier models exhibit little to no evidence of overfitting and that many models, even the most heavily overfit families, show strong signs of generalizable mathematical reasoning.
\section{Acknowledgements}
We would like to thank Dan Hendrycks, Adi Ganesh, Akilesh Praveen, Andrea Jaba, Will Zhou, Celia Chen, Apaar Shanker and Kamilė Lukošiūtė for their helpful comments and suggestions.

%% file: paper/appendix.tex
\clearpage
\appendix
\input{paper/checklist}

\section{Annotator Instructions}
\label{app:annotator_instructions}
We provide the annotator instructions given below.
\lstset{basicstyle=\fontfamily{fvm}\selectfont\footnotesize,breaklines=true}
\begin{lstlisting}[breaklines]
    
Welcome to the Grade School Math Question Development project. The goal of this project is to create questions and answers similar to what is found in an 8th-grade math quiz. Our goal is to develop high-quality questions that are almost the same as what is found in the dataset but are entirely unique. You will see three example questions and their corresponding answers in each task. These examples will guide you to create completely new questions and answers. It's important to note that you cannot use chatbots or language models to help you develop these Q&A pairs. You may be removed from the project if we detect any use of chatbots. Crucially, your Q&A pairs must be original creations and cannot be paraphrased versions of the examples.

Your workflow for this project will be as follows:

Review the examples: In each task you will be shown examples from an 8th-grade question-and-answer dataset. Review the examples to inform how you can create your question and answer pair.

Problem Creation: Problems should follow step guidance in the task. Don't reuse a problem setting. If you wrote a problem about Rogers trip to the grocery store, don't write another problem using the same premise. All questions should have a resolution of 1 or higher. We do not want any questions with a negative integer or zero as the answer. 

Craft the resolution steps: Calculations should be simple enough an 8th grader can complete with a pen and paper. Only use elementary arithmetic operations (addition, subtraction, multiplication, division)

Provide the final Answer: Answers should be a single integer value. Any units should be specified as part of the question (e.g. "How much money, in dollars, does Robert have?"). Simple decimal numbers (e.g. 3.25) can be part of the intermediate steps in the problem, but final answers should always be integers.

Check your work: We will utilize quality control process to ensure accuracy but it is crucial to check your work!
\end{lstlisting}

\begin{figure}[b!]
    \centering
    \includegraphics[width=1\linewidth]{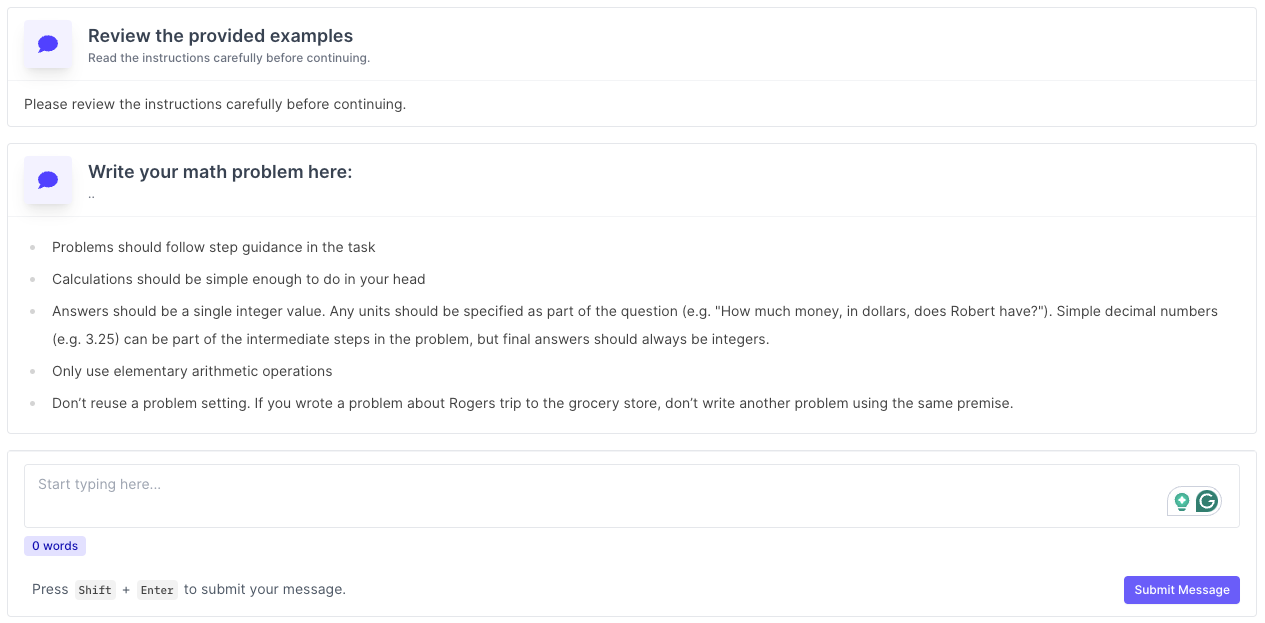}
    \caption{What annotators saw before seeing three example prompts drawn from GSM8k.}
    \label{fig:instructions}
\end{figure}

\clearpage

\section{N-shot Prompt (examples selected randomly from GSM8k train)}
Below is an example prompt. For each question, we select five random examples from GSM8k to use as n-shot examples, which vary for each new question from the GSM1k/GSM8k test set. While evaluation methods vary between models, this is the most common approach to evaluating GSM8k.
\label{app:prompt}
\begin{verbatim}
Question: Jen and Tyler are gymnasts practicing flips. Jen is practicing the triple-flip
while Tyler is practicing the double-flip. Jen did sixteen triple-flips during practice.
Tyler flipped in the air half the number of times Jen did. How many double-flips did Tyler do?
Answer: Jen did 16 triple-flips, so she did 16 * 3 = <<16*3=48>>48 flips.
Tyler did half the number of flips, so he did 48 / 2 = <<48/2=24>>24 flips.
A double flip has two flips, so Tyler did 24 / 2 = <<24/2=12>>12 double-flips.
#### 12
Question: Four people in a law firm are planning a party. Mary will buy a platter of pasta
for $20 and a loaf of bread for $2. Elle and Andrea will split the cost for buying 4 cans
of soda which cost $1.50 each, and chicken wings for $10. Joe will buy a cake that costs
$5. How much more will Mary spend than the rest of the firm put together?
Answer: Mary will spend $20 + $2 = $<<20+2=22>>22.
Elle and Andrea will spend $1.5 x 4 = $<<1.5*4=6>>6 for the soda.
Elle and Andrea will spend $6 + $10 = $<<6+10=16>>16 for the soda and chicken wings.
Elle, Andrea, and Joe together will spend $16 + $5 = $<<16+5=21>>21.
So, Mary will spend $22 - $21 = $<<22-21=1>>1 more than all of them combined.
#### 1
Question: A charcoal grill burns fifteen coals to ash every twenty minutes of grilling.
The grill ran for long enough to burn three bags of coals. Each bag of coal contains 60
coals. How long did the grill run?
Answer: The grill burned 3 * 60 = <<3*60=180>>180 coals.
It takes 20 minutes to burn 15 coals, so the grill ran for 180 / 15 * 20 =
<<180/15*20=240>>240 minutes.
#### 240
Question: A bear is preparing to hibernate for the winter and needs to gain 1000 pounds.
At the end of summer, the bear feasts on berries and small woodland animals. During autumn,
it devours acorns and salmon. It gained a fifth of the weight it needed from berries during
summer, and during autumn, it gained twice that amount from acorns. Salmon made up half of
the remaining weight it had needed to gain. How many pounds did it gain eating small animals?
Answer: The bear gained 1 / 5 * 1000 = <<1/5*1000=200>>200 pounds from berries.
It gained 2 * 200 = <<2*200=400>>400 pounds from acorns.
It still needed 1000 - 200 - 400 = <<1000-200-400=400>>400 pounds.
Thus, it gained 400 / 2 = <<400/2=200>>200 pounds from salmon.
Therefore, the bear gained 400 - 200 = <<400-200=200>>200 pounds from small animals.
#### 200
Question: Brendan can cut 8 yards of grass per day, he bought a lawnmower and it helped
him to cut more yards by Fifty percent per day. How many yards will Brendan be able to cut
after a week?
Answer: The additional yard Brendan can cut after buying the lawnmower is 8 x 0.50 =
<<8*0.50=4>>4 yards.
So, the total yards he can cut with the lawnmower is 8 + 4 = <<8+4=12>>12.
Therefore, the total number of yards he can cut in a week is 12 x 7 = <<12*7=84>>84 yards.
#### 84
\end{verbatim}
\clearpage

\section{Results with an Alternative Prompt}
\label{app:alternative}
As an ablation, we evaluate all models with an alternative prompt scheme and compare results with our primary findings. This prompt is available under the LM Evaluation Harness as a ``chain-of-thought'' prompt. However, manually examining the prompt (provided in full below) reveals that the primary difference with the standard n-shot prompt lies not in chain-of-thought reasoning but rather using a set of non-GSM8k problems as guiding examples as well as providing an alternative answer format. We choose to use the standard prompt to match typical evaluation methods widespread in the field but also report these results for completeness.
\begin{lstlisting}
Q: There are 15 trees in the grove. Grove workers will plant trees in the grove today. After they are done, there will be 21 trees. How many trees did the grove workers plant today?
A: There are 15 trees originally. Then there were 21 trees after some more were planted. So there must have been 21 - 15 = 6. The answer is 6.

Q: If there are 3 cars in the parking lot and 2 more cars arrive, how many cars are in the parking lot?
A: There are originally 3 cars. 2 more cars arrive. 3 + 2 = 5. The answer is 5.

Q: Leah had 32 chocolates and her sister had 42. If they ate 35, how many pieces do they have left in total?
A: Originally, Leah had 32 chocolates. Her sister had 42. So in total they had 32 + 42 = 74. After eating 35, they had 74 - 35 = 39. The answer is 39.

Q: Jason had 20 lollipops. He gave Denny some lollipops. Now Jason has 12 lollipops. How many lollipops did Jason give to Denny?
A: Jason started with 20 lollipops. Then he had 12 after giving some to Denny. So he gave Denny 20 - 12 = 8. The answer is 8.

Q: Shawn has five toys. For Christmas, he got two toys each from his mom and dad. How many toys does he have now?
A: Shawn started with 5 toys. If he got 2 toys each from his mom and dad, then that is 4 more toys. 5 + 4 = 9. The answer is 9.

Q: There were nine computers in the server room. Five more computers were installed each day, from monday to thursday. How many computers are now in the server room?
A: There were originally 9 computers. For each of 4 days, 5 more computers were added. So 5 * 4 = 20 computers were added. 9 + 20 is 29. The answer is 29.

Q: Michael had 58 golf balls. On tuesday, he lost 23 golf balls. On wednesday, he lost 2 more. How many golf balls did he have at the end of wednesday?
A: Michael started with 58 golf balls. After losing 23 on tuesday, he had 58 - 23 = 35. After losing 2 more, he had 35 - 2 = 33 golf balls. The answer is 33.

Q: Olivia has $23. She bought five bagels for $3 each. How much money does she have left?
A: Olivia had 23 dollars. 5 bagels for 3 dollars each will be 5 x 3 = 15 dollars. So she has 23 - 15 dollars left. 23 - 15 is 8. The answer is 8.
\end{lstlisting}

We report our results in Table~\ref{tab:gsm8k_0_0}. While there is significant variance based on prompt, the general trend of which model families are overfit is similar.

\begin{figure}[ht!]
    \centering
    \includegraphics[width=1\linewidth]{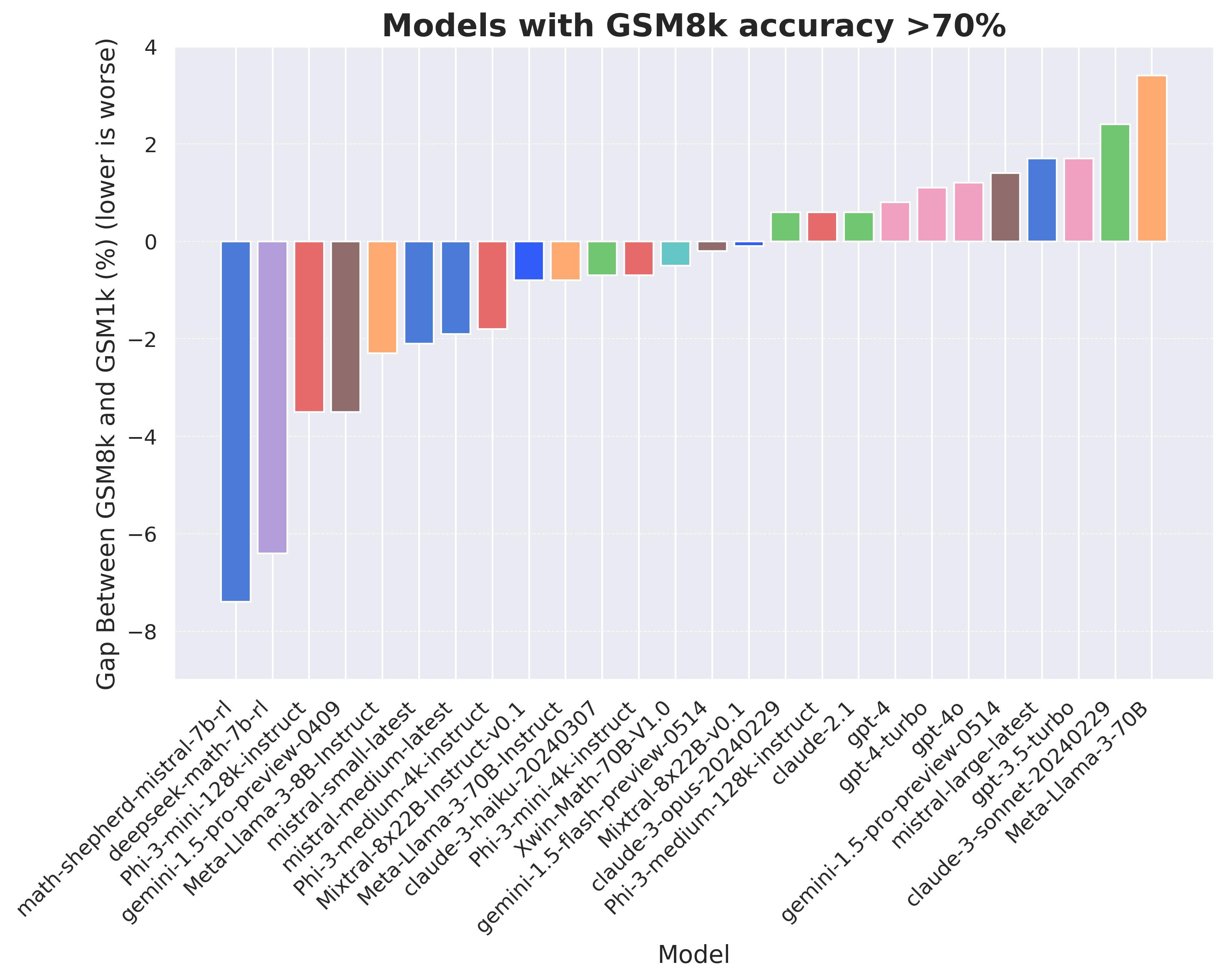}
    \caption{Gap in accuracy between GSM8k and GSM1k for models that score above 70\% on GSM8k.}
    \label{fig:topoverfit}
\end{figure}
\clearpage

\section{Results Table}
We report our full results in Table~\ref{tab:gsm8k_0_0}. Models are sorted by the difference in performance between GSM8k and GSM1k. Because all models are evaluated using the standard LM Evaluation Harness prompt and evaluation format, model performance on GSM8k may not match reported benchmark numbers. In particular, answers that do not match the 5-shot example format are marked incorrect even if they are otherwise ``correct.'' Our focus is primarily on the difference between GSM8k and GSM1k performance, holding evaluation setting constant. The Z-score and p-value are calculated for a two-tailed two proportion Z-test. Alternative prompt results are also included. For details, see Appendix~\ref{app:alternative}.
\label{app:table}
\input{paper/table.tex}
\clearpage
\input{paper/table_cot.tex}
\clearpage

\section{50 Examples from GSM1k}

A previous version of this paper mistakenly included some questions from a nonfinal version of GSM1k. A corrected table is below.

\label{app:50examples}
\input{paper/50examples.tex}
\clearpage

\section{Bar Chart of Performance Gaps Between GSM8k and GSM1k Across All Model Accuracies}

\begin{figure}[!ht]
\centering
\includegraphics[width=0.8\textwidth]{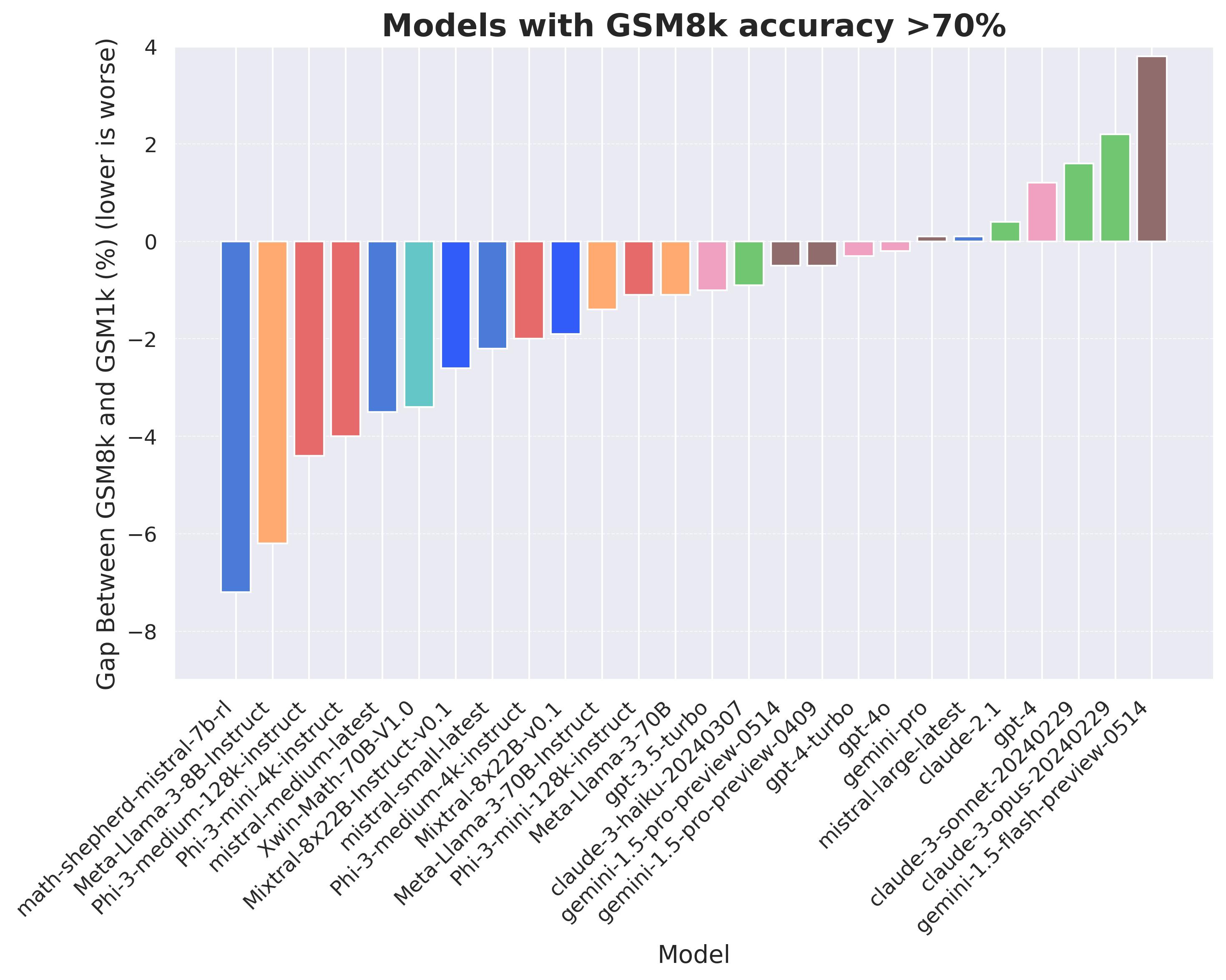}
\caption{Models with over 70\% accuracy on GSM8k. We observe that some models (e.g. Mistral, Phi) are overfit, while other models show little to no evidence of overfitting.}
\label{fig:goodaccuracy}
\end{figure}

\begin{figure}[!ht]
\centering
\includegraphics[width=0.8\textwidth]{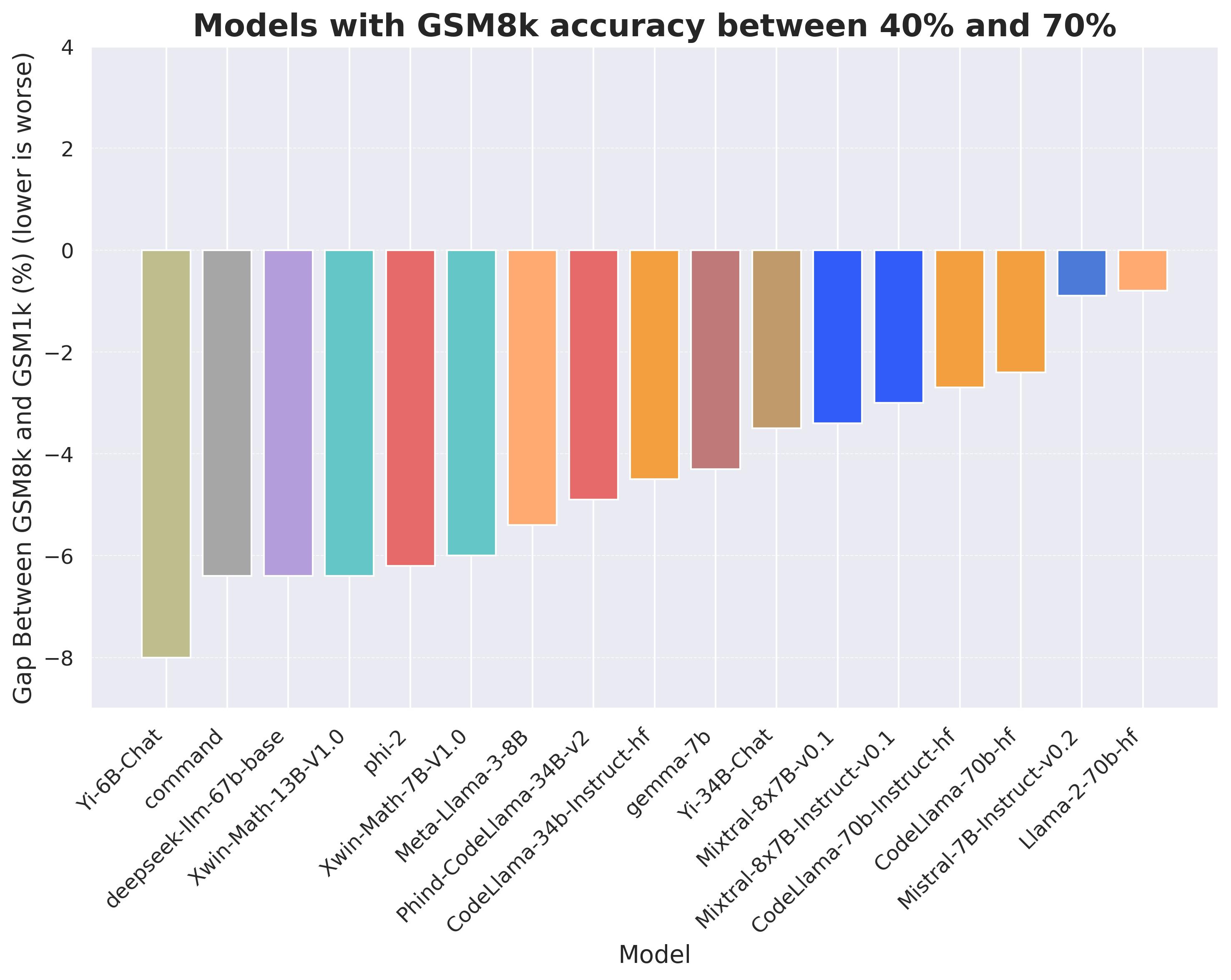}
\caption{Comparison of models with between 40 and 70\% accuracy on GSM8k. We observe that all models seem to fall below the line in this regime of model performance, though some models (e.g. Llama-2-70b) do much better than others.}
\label{fig:medaccuracy}
\end{figure}

\begin{figure}[!ht]
  \centering
  \includegraphics[width=0.8\linewidth]{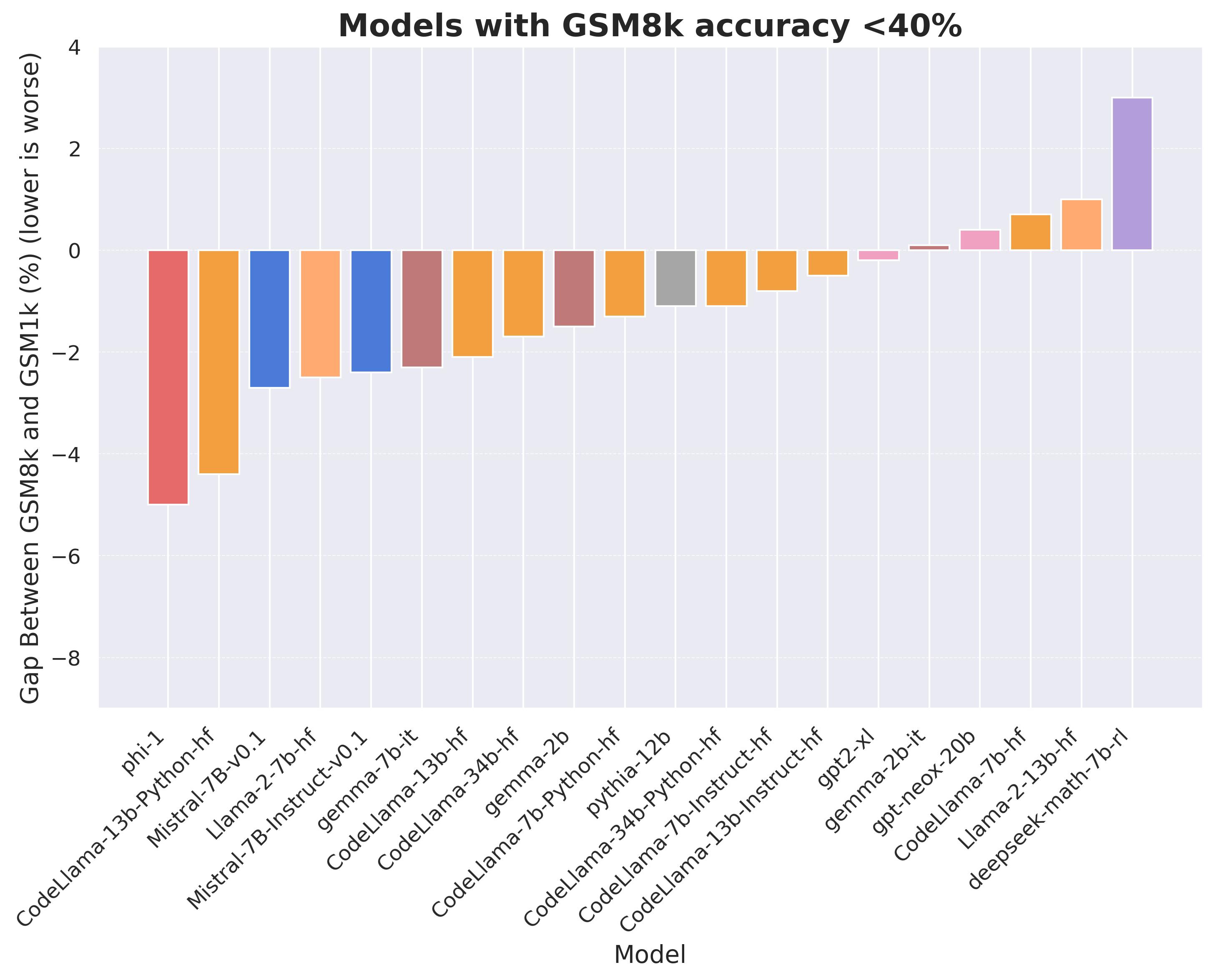}
  \caption{Models with less than 40\% accuracy on GSM8k.}
  \label{fig:badaccuracy}
\end{figure}

\clearpage
\section{Additional Plots}

\input{paper/xy_plots}

\begin{figure}[t!]
    \centering
    \includegraphics[width=1\linewidth]{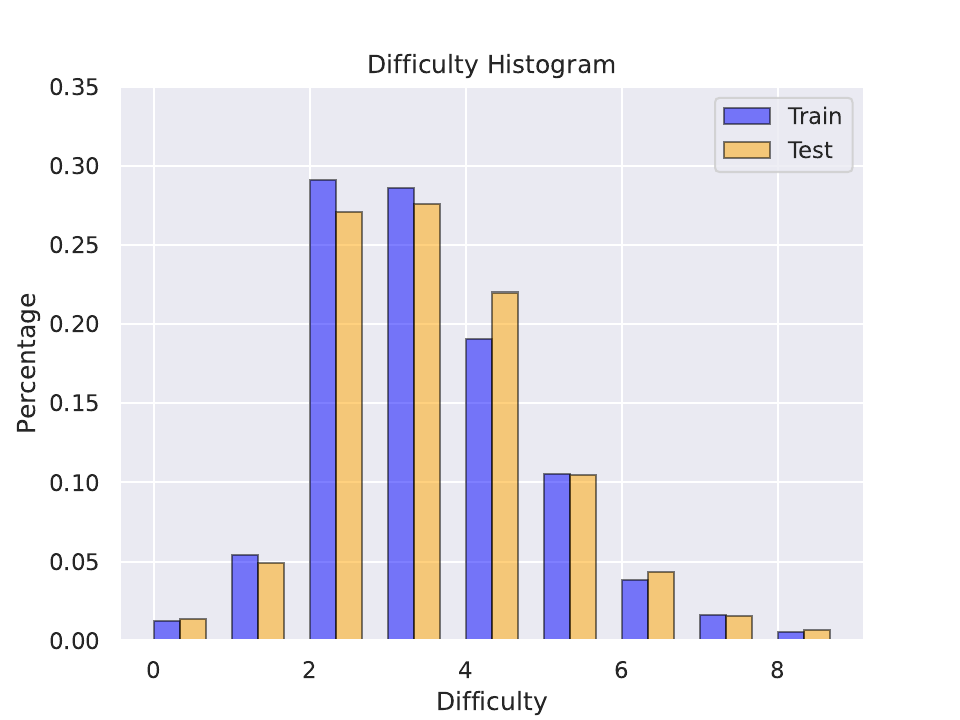}
    \caption{Approximate difficulty distribution of GSM8k train and test sets, measured by number of required steps to solve the problem. GSM1k annotators were instructed to create problems matching the overall distribution of the combined train and test difficulty distribution. The process of estimating problem difficulty is described in Section~\ref{sec:difficulty}.}
    \label{fig:instructions}
\end{figure}
\clearpage

\input{paper/human_extraction.tex}
\clearpage

\input{paper/katie_ablations}

\clearpage

\section{List of Models Evaluated Using Default HuggingFace}

We used vLLM to speed up model inference. A small number of models were not supported by vLLM at the time of initial evaluation, so we used the regular HuggingFace libraries to generate results. The list of these models is below.

\begin{itemize}
\item databricks/dbrx-base
\item databricks/dbrx-instruct
\item google/gemma-2b
\item google/gemma-7b
\item google/gemma-7b-it
\item google/gemma-2b-it
\item google/gemma-1.1-7b-it
\item google/gemma-1.1-2b-it
\item google/codegemma-7b
\item google/codegemma-7b-it
\item microsoft/Phi-3-mini-4k-instruct
\item microsoft/Phi-3-mini-128k-instruct
\item microsoft/Phi-3-medium-4k-instruct
\item microsoft/Phi-3-medium-128k-instruct
\end{itemize}

%% file: paper/checklist.tex
\section{Checklist}

\begin{enumerate}

\item For all authors...
\begin{enumerate}
  \item Do the main claims made in the abstract and introduction accurately reflect the paper's contributions and scope?
    \answerYes{}
  \item Did you describe the limitations of your work?
    \answerYes{}
  \item Did you discuss any potential negative societal impacts of your work?
    \answerNA{}
  \item Have you read the ethics review guidelines and ensured that your paper conforms to them?
    \answerYes{}
\end{enumerate}

\item If you are including theoretical results...
\begin{enumerate}
  \item Did you state the full set of assumptions of all theoretical results?
    \answerNA{}
	\item Did you include complete proofs of all theoretical results?
    \answerNA{}
\end{enumerate}

\item If you ran experiments (e.g. for benchmarks)...
\begin{enumerate}
  \item Did you include the code, data, and instructions needed to reproduce the main experimental results (either in the supplemental material or as a URL)?
    \answerYes{included in supplement, will open source shortly}
  \item Did you specify all the training details (e.g., data splits, hyperparameters, how they were chosen)?
    \answerNA{}
	\item Did you report error bars (e.g., with respect to the random seed after running experiments multiple times)?
    \answerYes{Appendix~\ref{app:table} includes standard errors and statistical significance.}
	\item Did you include the total amount of compute and the type of resources used (e.g., type of GPUs, internal cluster, or cloud provider)?
    \answerYes{We run all models on a cluster with 8 x A100 nodes. Most models complete evaluation within a few minutes.}
\end{enumerate}

\item If you are using existing assets (e.g., code, data, models) or curating/releasing new assets...
\begin{enumerate}
  \item If your work uses existing assets, did you cite the creators?
    \answerNA{}
  \item Did you mention the license of the assets?
    \answerNA{}
  \item Did you include any new assets either in the supplemental material or as a URL?
    \answerYes{GSM1k is yet unreleased, with a future release date. We provide the full dataset \href{https://docs.google.com/spreadsheets/d/1pCo9irpi-M3E9KmTevOVSLZi90CQqmgPj_JmR1UV5wc/edit?usp=sharing}{here}. We ask that reviewers refrain from sharing this dataset publicly.}
  \item Did you discuss whether and how consent was obtained from people whose data you're using/curating?
    \answerYes{All problems created by annotators hired by Scale AI.}
  \item Did you discuss whether the data you are using/curating contains personally identifiable information or offensive content?
    \answerNA{}
\end{enumerate}

\item If you used crowdsourcing or conducted research with human subjects...
\begin{enumerate}
  \item Did you include the full text of instructions given to participants and screenshots, if applicable?
    \answerYes{See Appendix~\ref{app:annotator_instructions}}
  \item Did you describe any potential participant risks, with links to Institutional Review Board (IRB) approvals, if applicable?
    \answerNA{}
  \item Did you include the estimated hourly wage paid to participants and the total amount spent on participant compensation?
    \answerYes{Annotators were paid 20-25 / hour, depending on performance, experience, and bonus incentives. In total, Scale paid out around 180K to human annotators to create this benchmark.}
\end{enumerate}

\end{enumerate}

\clearpage

\section{Dataset Documentation}

\begin{enumerate}
  \item \textbf{Construction.} GSM1k is a dataset of 1205 questions requiring elementary mathematical reasoning to solve. All problems are intended to be solvable using only the four basic arithmetic operators.
  \item \textbf{Creation.} GSM1k was created using human annotation from scratch without any usage of LLMs. Human annotators were hired by Scale AI and paid between 20 and 25 dollars per hour. All annotators were based in the United States. In total, this dataset paid out around \$180,000 dollars to human annotators, including costs resulting from problem creation and solving, quality assurance checks, as well as experiments done to compare the difficulty distribution with GSM8k.
  \item \textbf{Intent.} This dataset is intended to be used as a held-out version of GSM8k to measure data contamination. As such, it largely mimics the format and style of GSM8k. All answers are a non-negative integer.
  \item \textbf{Release.} Our dataset is not published at this time, to prevent risk of data contamination in future models. We will release Croissant metadata when the dataset is public, with the conditions described in the main paper.
  \item \textbf{Liability.} The authors bear all responsibility in case of violation of rights. Due to Scale AI commissioning the construction of this dataset from scratch primarily for this purpose of this paper, we do not anticipate any copyright or other issues. The dataset (yet unreleased) will be released with the MIT license. 
  \item \textbf{Preservation.} We plan to release the full dataset on Github as well as HuggingFace so it remains publicly accessible to anyone who wishes to use it. The formatting will be 1205 rows with a question and answer column.
\end{enumerate}

\clearpage

%% file: paper/table.tex
\begin{longtable}{p{0.35\textwidth}p{0.08\textwidth}p{0.08\textwidth}p{0.08\textwidth}p{0.10\textwidth}p{0.08\textwidth}}
\captionsetup{labelformat=empty}\caption{\textbf{\large{Standard Prompt}}} \\\toprule
\textbf{Model} & \textbf{Diff} & \textbf{GSM8k} & \textbf{GSM1k} & \textbf{Z-score} & \textbf{p-value} \\
\midrule
\endhead
\bottomrule
\endfoot
\texttt{Yi-6B-Chat} & 0.080 & 0.437 & 0.357 & 4.135 & 0.000 \\
\addlinespace
\texttt{math-shepherd-mistral-7b-rl} & 0.072 & 0.826 & 0.754 & 4.488 & 0.000 \\
\addlinespace
\texttt{command} & 0.065 & 0.447 & 0.383 & 3.336 & 0.000 \\
\addlinespace
\texttt{Xwin-Math-13B-V1.0} & 0.064 & 0.660 & 0.596 & 3.334 & 0.000 \\
\addlinespace
\texttt{phi-2} & 0.063 & 0.566 & 0.504 & 3.167 & 0.001 \\
\addlinespace
\texttt{Meta-Llama-3-8B-Instruct} & 0.062 & 0.752 & 0.690 & 3.532 & 0.000 \\
\addlinespace
\texttt{Xwin-Math-7B-V1.0} & 0.060 & 0.552 & 0.492 & 3.040 & 0.001 \\
\addlinespace
\texttt{Meta-Llama-3-8B} & 0.054 & 0.502 & 0.448 & 2.734 & 0.003 \\
\addlinespace
\texttt{phi-1.5} & 0.051 & 0.324 & 0.274 & 2.814 & 0.002 \\
\addlinespace
\texttt{Phind-CodeLlama-34B-v2} & 0.049 & 0.419 & 0.370 & 2.531 & 0.006 \\
\addlinespace
\texttt{CodeLlama-34b-Instruct-hf} & 0.045 & 0.426 & 0.381 & 2.338 & 0.010 \\
\addlinespace
\texttt{Phi-3-medium-128k-instruct} & 0.044 & 0.869 & 0.825 & 3.103 & 0.001 \\
\addlinespace
\texttt{CodeLlama-13b-Python-hf} & 0.044 & 0.223 & 0.179 & 2.759 & 0.003 \\
\addlinespace
\texttt{gemma-7b} & 0.043 & 0.519 & 0.476 & 2.198 & 0.014 \\
\addlinespace
\texttt{Phi-3-mini-4k-instruct} & 0.040 & 0.788 & 0.748 & 2.385 & 0.009 \\
\addlinespace
\texttt{Yi-34B-Chat} & 0.035 & 0.685 & 0.650 & 1.883 & 0.030 \\
\addlinespace
\texttt{mistral-medium-latest} & 0.035 & 0.790 & 0.755 & 2.104 & 0.018 \\
\addlinespace
\texttt{Mixtral-8x7B-v0.1} & 0.035 & 0.591 & 0.557 & 1.771 & 0.038 \\
\addlinespace
\texttt{Xwin-Math-70B-V1.0} & 0.034 & 0.806 & 0.772 & 2.107 & 0.018 \\
\addlinespace
\texttt{Mixtral-8x7B-Instruct-v0.1} & 0.030 & 0.660 & 0.630 & 1.588 & 0.056 \\
\addlinespace
\texttt{Mistral-7B-v0.1} & 0.027 & 0.391 & 0.364 & 1.421 & 0.078 \\
\addlinespace
\texttt{Mixtral-8x22B-Instruct-v0.1} & 0.026 & 0.872 & 0.846 & 1.913 & 0.028 \\
\addlinespace
\texttt{CodeLlama-70b-Instruct-hf} & 0.026 & 0.513 & 0.486 & 1.323 & 0.093 \\
\addlinespace
\texttt{Llama-2-7b-hf} & 0.025 & 0.141 & 0.116 & 1.892 & 0.029 \\
\addlinespace
\texttt{Mistral-7B-Instruct-v0.1} & 0.025 & 0.353 & 0.329 & 1.309 & 0.095 \\
\addlinespace
\texttt{CodeLlama-70b-hf} & 0.024 & 0.478 & 0.454 & 1.221 & 0.111 \\
\addlinespace
\texttt{gemma-7b-it} & 0.023 & 0.325 & 0.302 & 1.247 & 0.106 \\
\addlinespace
\texttt{mistral-small-latest} & 0.022 & 0.790 & 0.768 & 1.343 & 0.090 \\
\addlinespace
\texttt{CodeLlama-13b-hf} & 0.021 & 0.236 & 0.215 & 1.247 & 0.106 \\
\addlinespace
\texttt{Phi-3-medium-4k-instruct} & 0.020 & 0.874 & 0.854 & 1.519 & 0.064 \\
\addlinespace
\texttt{Mixtral-8x22B-v0.1} & 0.020 & 0.767 & 0.748 & 1.138 & 0.127 \\
\addlinespace
\texttt{CodeLlama-34b-hf} & 0.017 & 0.354 & 0.337 & 0.919 & 0.179 \\
\addlinespace
\texttt{gemma-2b} & 0.015 & 0.185 & 0.170 & 0.966 & 0.167 \\
\addlinespace
\texttt{Meta-Llama-3-70B-Instruct} & 0.014 & 0.914 & 0.900 & 1.251 & 0.105 \\
\addlinespace
\texttt{CodeLlama-7b-Python-hf} & 0.013 & 0.131 & 0.118 & 1.040 & 0.149 \\
\addlinespace
\texttt{dbrx-base} & 0.012 & 0.731 & 0.719 & 0.707 & 0.240 \\
\addlinespace
\texttt{pythia-12b} & 0.011 & 0.036 & 0.025 & 1.701 & 0.044 \\
\addlinespace
\texttt{Phi-3-mini-128k-instruct} & 0.011 & 0.757 & 0.746 & 0.645 & 0.260 \\
\addlinespace
\texttt{Meta-Llama-3-70B} & 0.011 & 0.817 & 0.806 & 0.707 & 0.240 \\
\addlinespace
\texttt{CodeLlama-34b-Python-hf} & 0.010 & 0.312 & 0.301 & 0.549 & 0.291 \\
\addlinespace
\texttt{gpt-3.5-turbo} & 0.009 & 0.760 & 0.750 & 0.546 & 0.293 \\
\addlinespace
\texttt{Mistral-7B-Instruct-v0.2} & 0.009 & 0.428 & 0.419 & 0.469 & 0.319 \\
\addlinespace
\texttt{claude-3-haiku-20240307} & 0.009 & 0.785 & 0.776 & 0.532 & 0.298 \\
\addlinespace
\texttt{Llama-2-70b-hf} & 0.008 & 0.552 & 0.544 & 0.445 & 0.328 \\
\addlinespace
\texttt{CodeLlama-7b-Instruct-hf} & 0.007 & 0.177 & 0.169 & 0.472 & 0.319 \\
\addlinespace
\texttt{gemini-1.5-pro-preview-0514} & 0.006 & 0.895 & 0.890 & 0.472 & 0.318 \\
\addlinespace
\texttt{gemini-1.5-pro-preview-0409} & 0.005 & 0.897 & 0.892 & 0.403 & 0.343 \\
\addlinespace
\texttt{CodeLlama-13b-Instruct-hf} & 0.005 & 0.267 & 0.262 & 0.257 & 0.399 \\
\addlinespace
\texttt{dbrx-instruct} & 0.004 & 0.730 & 0.726 & 0.211 & 0.417 \\
\addlinespace
\texttt{gpt-4-turbo} & 0.003 & 0.898 & 0.895 & 0.270 & 0.394 \\
\addlinespace
\texttt{gpt2-xl} & 0.002 & 0.009 & 0.007 & 0.778 & 0.218 \\
\addlinespace
\texttt{gpt-4o} & 0.002 & 0.931 & 0.929 & 0.219 & 0.413 \\
\addlinespace
\texttt{gemini-pro} & -0.001 & 0.792 & 0.793 & -0.081 & 0.532 \\
\addlinespace
\texttt{mistral-large-latest} & -0.001 & 0.853 & 0.854 & -0.049 & 0.519 \\
\addlinespace
\texttt{gemma-2b-it} & -0.001 & 0.111 & 0.112 & -0.106 & 0.542 \\
\addlinespace
\texttt{claude-2.1} & -0.004 & 0.887 & 0.891 & -0.336 & 0.632 \\
\addlinespace
\texttt{CodeLlama-7b-hf} & -0.007 & 0.126 & 0.133 & -0.525 & 0.700 \\
\addlinespace
\texttt{Llama-2-13b-hf} & -0.011 & 0.236 & 0.246 & -0.629 & 0.735 \\
\addlinespace
\texttt{gpt-4} & -0.012 & 0.911 & 0.923 & -1.161 & 0.877 \\
\addlinespace
\texttt{claude-3-sonnet-20240229} & -0.016 & 0.719 & 0.735 & -0.894 & 0.814 \\
\addlinespace
\texttt{claude-3-opus-20240229} & -0.022 & 0.802 & 0.824 & -1.421 & 0.922 \\
\addlinespace
\texttt{deepseek-math-7b-rl} & -0.031 & 0.187 & 0.217 & -1.963 & 0.975 \\
\addlinespace
\texttt{gemini-1.5-flash-preview-0514} & -0.038 & 0.797 & 0.835 & -2.507 & 0.994 \\
\end{longtable}
\label{tab:gsm8k_0_0}

%% file: paper/table_cot.tex
\begin{longtable}{p{0.35\textwidth}p{0.08\textwidth}p{0.08\textwidth}p{0.08\textwidth}p{0.10\textwidth}p{0.08\textwidth}}
\captionsetup{labelformat=empty}\caption{\textbf{\large{Alternative Prompt}}} \\\toprule
\textbf{Model} & \textbf{Diff} & \textbf{GSM8k} & \textbf{GSM1k} & \textbf{Z-score} & \textbf{p-value} \\
\midrule
\endhead
\bottomrule
\endfoot
\texttt{math-shepherd-mistral-7b-rl} & 0.074 & 0.820 & 0.746 & 4.504 & 0.000 \\
\addlinespace
\texttt{deepseek-math-7b-rl} & 0.064 & 0.760 & 0.696 & 3.672 & 0.000 \\
\addlinespace
\texttt{Yi-6B-Chat} & 0.058 & 0.426 & 0.368 & 2.964 & 0.002 \\
\addlinespace
\texttt{CodeLlama-34b-Python-hf} & 0.056 & 0.337 & 0.280 & 3.059 & 0.001 \\
\addlinespace
\texttt{command} & 0.051 & 0.457 & 0.407 & 2.596 & 0.005 \\
\addlinespace
\texttt{phi-1.5} & 0.050 & 0.321 & 0.271 & 2.786 & 0.003 \\
\addlinespace
\texttt{Xwin-Math-13B-V1.0} & 0.042 & 0.662 & 0.620 & 2.212 & 0.013 \\
\addlinespace
\texttt{CodeLlama-70b-Instruct-hf} & 0.040 & 0.529 & 0.489 & 2.047 & 0.020 \\
\addlinespace
\texttt{CodeLlama-70b-hf} & 0.037 & 0.517 & 0.480 & 1.878 & 0.030 \\
\addlinespace
\texttt{gemma-7b} & 0.036 & 0.568 & 0.532 & 1.826 & 0.034 \\
\addlinespace
\texttt{gemini-1.5-pro-preview-0409} & 0.035 & 0.908 & 0.873 & 2.883 & 0.002 \\
\addlinespace
\texttt{Phi-3-mini-128k-instruct} & 0.035 & 0.818 & 0.783 & 2.211 & 0.014 \\
\addlinespace
\texttt{phi-2} & 0.034 & 0.552 & 0.518 & 1.744 & 0.041 \\
\addlinespace
\texttt{Xwin-Math-7B-V1.0} & 0.033 & 0.530 & 0.497 & 1.680 & 0.046 \\
\addlinespace
\texttt{CodeLlama-7b-hf} & 0.027 & 0.123 & 0.095 & 2.242 & 0.012 \\
\addlinespace
\texttt{Mistral-7B-Instruct-v0.2} & 0.027 & 0.437 & 0.410 & 1.389 & 0.082 \\
\addlinespace
\texttt{dbrx-base} & 0.024 & 0.712 & 0.688 & 1.322 & 0.093 \\
\addlinespace
\texttt{gemma-7b-it} & 0.023 & 0.254 & 0.231 & 1.394 & 0.082 \\
\addlinespace
\texttt{Meta-Llama-3-8B-Instruct} & 0.023 & 0.774 & 0.751 & 1.362 & 0.087 \\
\addlinespace
\texttt{CodeLlama-7b-Instruct-hf} & 0.022 & 0.187 & 0.165 & 1.493 & 0.068 \\
\addlinespace
\texttt{Yi-34B-Chat} & 0.022 & 0.679 & 0.656 & 1.170 & 0.121 \\
\addlinespace
\texttt{mistral-small-latest} & 0.021 & 0.782 & 0.761 & 1.305 & 0.096 \\
\addlinespace
\texttt{CodeLlama-13b-Python-hf} & 0.021 & 0.218 & 0.197 & 1.299 & 0.097 \\
\addlinespace
\texttt{CodeLlama-34b-hf} & 0.020 & 0.330 & 0.310 & 1.097 & 0.136 \\
\addlinespace
\texttt{pythia-12b} & 0.019 & 0.049 & 0.030 & 2.553 & 0.005 \\
\addlinespace
\texttt{mistral-medium-latest} & 0.019 & 0.789 & 0.770 & 1.152 & 0.125 \\
\addlinespace
\texttt{Phi-3-medium-4k-instruct} & 0.018 & 0.901 & 0.883 & 1.428 & 0.077 \\
\addlinespace
\texttt{dbrx-instruct} & 0.016 & 0.713 & 0.697 & 0.924 & 0.178 \\
\addlinespace
\texttt{Mixtral-8x7B-Instruct-v0.1} & 0.016 & 0.679 & 0.662 & 0.870 & 0.192 \\
\addlinespace
\texttt{gemma-2b} & 0.016 & 0.194 & 0.178 & 1.020 & 0.154 \\
\addlinespace
\texttt{Llama-2-7b-hf} & 0.014 & 0.142 & 0.128 & 1.021 & 0.154 \\
\addlinespace
\texttt{Phind-CodeLlama-34B-v2} & 0.014 & 0.398 & 0.384 & 0.728 & 0.233 \\
\addlinespace
\texttt{Mixtral-8x7B-v0.1} & 0.013 & 0.614 & 0.601 & 0.690 & 0.245 \\
\addlinespace
\texttt{Meta-Llama-3-8B} & 0.013 & 0.547 & 0.534 & 0.660 & 0.255 \\
\addlinespace
\texttt{gemini-pro} & 0.012 & 0.688 & 0.676 & 0.677 & 0.249 \\
\addlinespace
\texttt{Mistral-7B-v0.1} & 0.011 & 0.431 & 0.420 & 0.583 & 0.280 \\
\addlinespace
\texttt{Meta-Llama-3-70B-Instruct} & 0.008 & 0.907 & 0.899 & 0.714 & 0.238 \\
\addlinespace
\texttt{Mixtral-8x22B-Instruct-v0.1} & 0.008 & 0.890 & 0.882 & 0.612 & 0.270 \\
\addlinespace
\texttt{Phi-3-mini-4k-instruct} & 0.007 & 0.807 & 0.800 & 0.474 & 0.318 \\
\addlinespace
\texttt{claude-3-haiku-20240307} & 0.006 & 0.792 & 0.785 & 0.416 & 0.339 \\
\addlinespace
\texttt{Llama-2-13b-hf} & 0.005 & 0.281 & 0.276 & 0.298 & 0.383 \\
\addlinespace
\texttt{Xwin-Math-70B-V1.0} & 0.005 & 0.808 & 0.803 & 0.319 & 0.375 \\
\addlinespace
\texttt{gpt2-xl} & 0.004 & 0.006 & 0.002 & 1.422 & 0.078 \\
\addlinespace
\texttt{Mixtral-8x22B-v0.1} & 0.002 & 0.808 & 0.807 & 0.115 & 0.454 \\
\addlinespace
\texttt{gemini-1.5-flash-preview-0514} & 0.001 & 0.810 & 0.808 & 0.110 & 0.456 \\
\addlinespace
\texttt{CodeLlama-7b-Python-hf} & 0.001 & 0.119 & 0.118 & 0.112 & 0.455 \\
\addlinespace
\texttt{CodeLlama-13b-Instruct-hf} & -0.000 & 0.284 & 0.285 & -0.028 & 0.511 \\
\addlinespace
\texttt{gemma-2b-it} & -0.000 & 0.101 & 0.101 & -0.064 & 0.526 \\
\addlinespace
\texttt{CodeLlama-34b-Instruct-hf} & -0.002 & 0.403 & 0.404 & -0.073 & 0.529 \\
\addlinespace
\texttt{CodeLlama-13b-hf} & -0.004 & 0.213 & 0.217 & -0.232 & 0.592 \\
\addlinespace
\texttt{Phi-3-medium-128k-instruct} & -0.005 & 0.870 & 0.876 & -0.368 & 0.644 \\
\addlinespace
\texttt{claude-3-opus-20240229} & -0.006 & 0.830 & 0.836 & -0.396 & 0.654 \\
\addlinespace
\texttt{claude-2.1} & -0.006 & 0.836 & 0.842 & -0.425 & 0.665 \\
\addlinespace
\texttt{gpt-4} & -0.008 & 0.919 & 0.927 & -0.790 & 0.785 \\
\addlinespace
\texttt{gpt-4-turbo} & -0.011 & 0.847 & 0.858 & -0.825 & 0.795 \\
\addlinespace
\texttt{Mistral-7B-Instruct-v0.1} & -0.011 & 0.340 & 0.352 & -0.617 & 0.731 \\
\addlinespace
\texttt{gpt-4o} & -0.012 & 0.913 & 0.925 & -1.188 & 0.882 \\
\addlinespace
\texttt{Llama-2-70b-hf} & -0.013 & 0.572 & 0.585 & -0.636 & 0.738 \\
\addlinespace
\texttt{gemini-1.5-pro-preview-0514} & -0.014 & 0.802 & 0.816 & -0.894 & 0.814 \\
\addlinespace
\texttt{mistral-large-latest} & -0.017 & 0.854 & 0.871 & -1.228 & 0.890 \\
\addlinespace
\texttt{gpt-3.5-turbo} & -0.017 & 0.742 & 0.759 & -0.994 & 0.840 \\
\addlinespace
\texttt{claude-3-sonnet-20240229} & -0.024 & 0.713 & 0.737 & -1.326 & 0.908 \\
\addlinespace
\texttt{Meta-Llama-3-70B} & -0.034 & 0.815 & 0.849 & -2.287 & 0.989 \\
\end{longtable}
\label{tab:gsm8k_alt_0_0}

%% file: paper/50examples.tex
\begin{longtable}{|>{\centering\arraybackslash}m{0.1\textwidth}|@{\hspace{5mm}}m{0.75\textwidth}@{\hspace{5mm}}|>{\centering\arraybackslash}m{0.1\textwidth}|}
\hline
\textbf{No.} & \textbf{Question} & \textbf{Answer} \\
\hline
\endhead
\hline
\endfoot
1 & \rule{0pt}{2.5ex}Gabriela has \$65.00 and is shopping for groceries so that her grandmother can make her favorite kale soup. She needs heavy cream, kale, cauliflower, and meat (bacon and sausage). Gabriella spends 40\% of her money on the meat. She spends \$5.00 less than one-third of the remaining money on heavy cream. Cauliflower costs three-fourth of the price of the heavy cream and the kale costs \$2.00 less than the cauliflower. As Gabriela leaves the store, she spends one-third of her remaining money on her grandmother's favorite Girl Scout Cookies. How much money, in dollars, does Gabriela spend on Girl Scout cookies?\rule{0pt}{2.5ex} & 7 \\
\hline
2 & \rule{0pt}{2.5ex}Bernie is a street performer who plays guitar. On average, he breaks three guitar strings a week, and each guitar string costs \$3 to replace. How much does he spend on guitar strings over the course of an entire year?\rule{0pt}{2.5ex} & 468 \\
\hline
3 & \rule{0pt}{2.5ex}John Henry is competing against a machine to see who can dig a tunnel more quickly. John works without rest, and excavates at a rate of 6 cubic feet of rock per hour. The machine excavates more quickly but needs to be refueled and maintained by its operator for 30 minutes out of every hour. When it's not under maintenance, the machine excavates at a rate of 10 cubic feet of stone per hour. Provided that the competition lasts for 8 hours, how much more rock will John have excavated compared to the machine?\rule{0pt}{2.5ex} & 8 \\
\hline
4 & \rule{0pt}{2.5ex}Colin is playing dice with his friend Eoin and needs some help keeping track of his score. He begins with 5 points and wins 6 points in the first round. In the second round, he won twice as many points as he won in the first round. In the third round, he had a fantastic roll and was able to triple his total point count! How many points did Colin end the game with?\rule{0pt}{2.5ex} & 69 \\
\hline
5 & \rule{0pt}{2.5ex}Marge got a job so she can buy her first car. Her job pays \$15/hr and she works there 30 hours a week. The car Marge wants is \$3600. How many weeks does Marge need to work to buy the car?\rule{0pt}{2.5ex} & 8 \\
\hline
6 & \rule{0pt}{2.5ex}Andy’s soccer team needs 80 points to finish in first place. His team plays 38 games, and he gets 3 points for each win, 1 point for each tie, and 0 points for each loss. After 26 games, the team has 15 wins, 5 ties, and 6 losses. How many more points does Andy’s team need to reach 80 points?\rule{0pt}{2.5ex} & 30 \\
\hline
7 & \rule{0pt}{2.5ex}Molly wants to win the contest at school for reading 25 books before the end of May. So far, she has read 5 books by the end of January. How many more books will she need to read on average each month until the end of May to win the contest?\rule{0pt}{2.5ex} & 5 \\
\hline
8 & \rule{0pt}{2.5ex}Ms. Crabapple has a bag of jelly beans that she is going to divide equally among all of her 32 students who complete their homework every day over the course of a week. The bag has 384 jellybeans in it. Unfortunately, many of Ms. Crabapple's students have a poorly developed work ethic, and only half of them complete all of the required homework. How many jelly beans will each of the eligible students receive?\rule{0pt}{2.5ex} & 24 \\
\hline
9 & \rule{0pt}{2.5ex}Bob has to read 2 books and 3 articles, while Emily has to read 4 books and 2 articles. Each book has 3 chapters and each chapter has 4 paragraphs. Each article has 4 sections and each section has 2 paragraphs. How many paragraphs in total will Bob and Emily read?\rule{0pt}{2.5ex} & 112 \\
\hline
10 & \rule{0pt}{2.5ex}Leah and 2 of her friends go to an all-you-can-eat dumpling buffet. Leah's 1st friend ate 30 dumplings, her 2nd friend ate twice as many dumplings as her 1st friend, and Leah ate 1.5 times as many dumplings as her 2nd friend. How many dumplings in total did Leah and her friends eat?\rule{0pt}{2.5ex} & 180 \\
\hline
11 & \rule{0pt}{2.5ex}Francis has a bowl of candy in front of him. There are three different flavors of candies that he’s eaten over the course of 3 hours. He’s eaten ten lemon, four orange, and sixteen cherry-flavored candies. If there were twenty of each when he started, how much of an average percentage is still left?\rule{0pt}{2.5ex} & 50 \\
\hline
12 & \rule{0pt}{2.5ex}Maryann is saving up for a new bike that costs \$450. She already has \$120 saved up. She earns \$15 per hour at her part-time job. How many hours does she need to work to afford the bike?\rule{0pt}{2.5ex} & 22 \\
\hline
13 & \rule{0pt}{2.5ex}Henry is renovating his kitchen and adding a new tile floor. He needs to cover an area of 200 square feet. He has a stack of tiles that measure 0.5 feet in length and width. He can get 40 tiles done per hour. Henry works for 6 hours at that rate, then has some coffee and works at a faster rate for the next 2 hours (60 tiles per hour). Henry runs out of tiles, so he goes to a store to purchase the remaining tiles needed to finish the floor. Given that the price per tile is \$2.50, how much will he need to spend at the store to get exactly enough tiles to finish the floor?\rule{0pt}{2.5ex} & 1100 \\
\hline
14 & \rule{0pt}{2.5ex}A painter needs to paint 3 houses. The first house requires 14 gallons of paint, the second house requires twice as much paint as the first, and the third house needs half as much paint as the second house. If one gallon of paint costs \$35 and the painter gets a bulk discount of 10\% for purchases over 30 gallons, how much will the paint cost in total?\rule{0pt}{2.5ex} & 1764 \\
\hline
15 & \rule{0pt}{2.5ex}A coal miner is loading up coal into mine carts. During the first hour of the day, he is able to load 15 carts. His boss yells at him after that, so for each of the next three hours, he loads twice as many carts. Each cart weighs 78 pounds. What was the total weight of the coal he loaded on this day?\rule{0pt}{2.5ex} & 8190 \\
\hline
16 & \rule{0pt}{2.5ex}A plane owned by Sunny Skies Airlines is flying from Indianapolis to Phoenix. The plane holds 180 passengers and is 2/3 full. Each passenger brings 2 carry-on bags and is charged a carry-on bag fee of \$35 per bag. How much money does Sunny Skies Airlines collect for the carry-on bag fees for this flight?\rule{0pt}{2.5ex} & 8400 \\
\hline
17 & \rule{0pt}{2.5ex}Sally went to the mall to buy clothes for the summer. She went to Forever 21 and bought 4 tops, each had different prices, \$12.99, \$6.99, \$17.99, \$21.99, and 3 pants each priced at \$15.99. If her subtotal is over \$75, she gets a discount of 15\% on her purchase at that store. Then she goes to Shoe Palace and buys 2 shoes for a total of \$123.26. How much money did Sally spend at the mall?\rule{0pt}{2.5ex} & 215 \\
\hline
18 & \rule{0pt}{2.5ex}Dean wants to buy flowers to make arrangements for a party. He is going to make 12 arrangements. He wants to include 4 roses and 3 daisies in each arrangement. Roses come by the dozens and are \$15 for each dozen. Daisies come in groups of 4 and are \$8 for the set. How much will it cost for Dean to make all 12 arrangements?\rule{0pt}{2.5ex} & 132 \\
\hline
19 & \rule{0pt}{2.5ex}Alex plans to adopt a new cat and needs help planning a budget for this event. The adoption fee is \$200, and it includes all the essential veterinary care needed for a kitten, but she also needs to buy other supplies for the cat when she brings it home. The litter boxes cost \$30, one package of litter costs \$17, a bag of dry food costs \$55, and the wet food costs \$1.50 per can. Alex will buy 2 litter boxes, 3 packages of litter, one bag of dry food, and 12 cans of wet food. How much money should Alex make sure she has before beginning the process of adopting her new cat?\rule{0pt}{2.5ex} & 384 \\
\hline
20 & \rule{0pt}{2.5ex}Samantha is saving money for a new bike by doing chores. She earns \$5 for every chore she completes. If she does 3 chores each day for a week, and then uses \$25 to buy a helmet, how much money does she have left at the end of the week?\rule{0pt}{2.5ex} & 80 \\
\hline
21 & \rule{0pt}{2.5ex}Frank sneaks out before his break at 3:20 pm and gets back at 4:05. If his break was only supposed to be half an hour, for how much longer did Frank sneak out?\rule{0pt}{2.5ex} & 15 \\
\hline
22 & \rule{0pt}{2.5ex}Janet wants to listen to 20 music albums by the end of the week. If she just finished her twelfth album and today is Thursday, how many albums per day would she have to listen to by Saturday?\rule{0pt}{2.5ex} & 4 \\
\hline
23 & \rule{0pt}{2.5ex}Hana wants to donate her clothes to a local charity. After going through her closet she ended up with 2 boxes of pants, 3 boxes of dresses, 1 box of shoes, and boxes of shirts. The number of boxes with shirts was 3 more than the other three boxes combined. How many boxes of shirts does she have to donate?\rule{0pt}{2.5ex} & 9 \\
\hline
24 & \rule{0pt}{2.5ex}Gayle has a lawnmowing business. Lawn 1 takes 15 minutes to mow. Lawn 2 takes 18 more minutes than Lawn 1. Lawn 3 takes 20\% more time to mow than Lawn 1. She is paid \$2.50 per minute for the time she spends. However, she gives her customers a 20\% discount. How much money does she make from mowing all three lawns?\rule{0pt}{2.5ex} & 132 \\
\hline
25 & \rule{0pt}{2.5ex}Frank ordered a whole chicken, 6 cans of chopped chicken breast, 1 lb. of macadamia nuts, and 4 bags of frozen broccoli. Each item has the following respective prices: \$12 per chicken, \$2 per can, \$24/lb., \$3 per bag. The sales tax was 10\% of the total cost and the tip was half the price of the whole chicken. How much did Frank pay for his order?\rule{0pt}{2.5ex} & 72 \\
\hline
26 & \rule{0pt}{2.5ex}Milo can bench press half as much weight as Doug can squat, and Doug can squat twice as much weight as Diane can squat. If Diana squats 125 pounds, how much weight can Milo bench press?\rule{0pt}{2.5ex} & 125 \\
\hline
27 & \rule{0pt}{2.5ex}Pablo is trying to make breakfast for his family. His wife eats 4 pancakes. His son eats 2 pancakes. Pablo wants to eat 4 pancakes. One box of pancake mix will make 5 pancakes. How many boxes of pancake mix will he need?\rule{0pt}{2.5ex} & 2 \\
\hline
28 & \rule{0pt}{2.5ex}Jim wants to spend 15\% of his monthly earnings on groceries. He makes \$2500/month. How much money will he have left over?\rule{0pt}{2.5ex} & 2125 \\
\hline
29 & \rule{0pt}{2.5ex}A school is ordering tablets and laptops for three classrooms. Each classroom will receive 4 tablets and 3 laptops. If each tablet costs \$250 and each laptop costs \$600, how much will the school spend in total for all three classrooms?\rule{0pt}{2.5ex} & 8400 \\
\hline
30 & \rule{0pt}{2.5ex} Grant takes 3 minutes to put on his pajamas. He brushes his teeth for 2 minutes. Then, he washes his face and brushes his hair for another 2 minutes. Finally, he reads a book for a while and turns off the light for bed. If Grant begins his routine at 8:15 pm and turns off the lights at 8:47 pm, for how long does Grant read a book?\rule{0pt}{2.5ex} & 25 \\
\hline
31 & \rule{0pt}{2.5ex}Bellemere owns a tangerine orchard with 50 trees. Each tree produces 80 tangerines. She wants to sell 600 tangerines at her local farmer's market. If she picks the same amount of tangerines from every tree, how many tangerines will be left on each tree?\rule{0pt}{2.5ex} & 68 \\
\hline
32 & \rule{0pt}{2.5ex}A charity puts out a telethon for a cause. Within 15 minutes, seventy-seven people donated \$3 each, and 231 people donated four dollars each. How much does the charity receive within this time?\rule{0pt}{2.5ex} & 1155 \\
\hline
33 & \rule{0pt}{2.5ex}A school is selling baskets for a fundraiser. There are three baskets containing the following items: * Blue basket: a ball, cup, and notebook. * Red basket: a cup, bell, and hat. * Green basket: a hat, pen, and notebook. The costs of the items in the baskets are as follows: * \$1: ball, notebook, and pen * \$2: cup, bell, and hat Jane buys 6 red baskets and 5 blue baskets. Jim buys 3 red baskets and 2 green baskets. Since they purchase so many, they receive a discount. Jane gets an \$8 discount and Jim also gets a \$2 discount. How many times more does Jane spend than Jim?\rule{0pt}{2.5ex} & 2 \\
\hline
34 & \rule{0pt}{2.5ex}Mr. Gordon has 14 boys in his first period class which is twice the number of girls in class. Two of the girls in class have blonde hair and the rest have brown hair. How many girls with brown hair are in his class?\rule{0pt}{2.5ex} & 5 \\
\hline
35 & \rule{0pt}{2.5ex}Albert gets paid \$15 an hour. He gets time and a half if he works over forty hours a week. Last week, he worked 48 hours. He plans to do this two weeks in a row. How much money will he be paid in overtime for those two weeks?\rule{0pt}{2.5ex} & 360 \\
\hline
36 & \rule{0pt}{2.5ex}Beth, Anna, and Kim went to a book fair. Beth had two books less than Anna while Kim had four more books than Anna. Beth had \$20 with her and was now left with \$8. If all books are priced at \$4, how much, in dollars, did Kim spend on her books?\rule{0pt}{2.5ex} & 36 \\
\hline
37 & \rule{0pt}{2.5ex}4 friends are going on a road trip. Their names are Alex, Bethany, Carlos, and Drew. They drive at a rate of 65, 75, 60, and 50 mph, respectively. Alex drives for 2 hours, Bethany for 4, and Carlos and Drew each drive for 3 hours. They are using a car with a fuel efficiency of 20 miles per gallon of gas. If, along their route, gas costs \$3 per gallon, how much money (in dollars) will they need to spend on gas? Assume they begin their journey at a gas station with an empty tank of gas.\rule{0pt}{2.5ex} & 114 \\
\hline
38 & \rule{0pt}{2.5ex}The Genco Olive Oil Company has received ninety-nine orders for ninety-nine barrels of olive oil each. Out of those shipped, 33 orders were sent back due to clerical or product errors. How many total barrels of olive oil were not returned?\rule{0pt}{2.5ex} & 6534 \\
\hline
39 & \rule{0pt}{2.5ex}There is a very large room that has 4 tables, 1 sofa and 2 chairs that have 4 legs each. There are also 3 tables with 3 legs each, 1 table with 1 leg, and 1 rocking chair with 2 legs. How many legs of tables are there in the room?\rule{0pt}{2.5ex} & 26 \\
\hline
40 & \rule{0pt}{2.5ex}A classroom has 24 students, and the teacher has arranged a field trip. If the cost per student for the trip is \$15 and the teacher already has \$120 from a class fund, how many more dollars does the teacher need to cover the total cost of the trip for all students?\rule{0pt}{2.5ex} & 240 \\
\hline
41 & \rule{0pt}{2.5ex} Rachel and Shauna go out to dinner. Dinner costs \$68.25 in total (without taxes). Rachel's meal costs 1/3 of the total price, while Shauna's meal costs 2/3 of the total price. How much did Shauna's meal cost (round to the nearest dollar)?\rule{0pt}{2.5ex} & 46 \\
\hline
42 & \rule{0pt}{2.5ex}Olivia owns a local hotel and needs to drive up business. She is planning to give a special deal to anyone who signs up for a membership card. Her idea is to give them 20\% off their first night and 10\% off on every night they stay after that. If her first new customer pays \$616 for their stay, and each night costs \$140 before discounts, how many nights did they stay at the hotel?\rule{0pt}{2.5ex} & 5 \\
\hline
43 & \rule{0pt}{2.5ex}Johnny has 8 green balls. He has five fewer than twice that in red balls. How many total balls does Johnny have?\rule{0pt}{2.5ex} & 19 \\
\hline
44 & \rule{0pt}{2.5ex}30 students are in a class. 1/5 of them are 12 years old, 1/3 are 13 years old. 1/10 of them are 11 years old. How many of them are not 11, 12, or 13 years old?\rule{0pt}{2.5ex} & 11 \\
\hline
45 & \rule{0pt}{2.5ex}Francis loves sandwiches. He gets his usual from his favorite deli: two “Big Boy” sandwiches, and a glass-bottled soda. A “Big Boy” costs \$15.25 and the soda costs \$3.75. His friend Lars calls him and asks for a double-sweet soda that’s \$4.75. If Francis pays all of this with \$40 and asks for his change back in only quarters, how many quarters will he get?\rule{0pt}{2.5ex} & 4 \\
\hline
46 & \rule{0pt}{2.5ex}A factory needs to produce 960 pieces of toy boats. They are only able to produce 1/6th of their goal a day. 5 toy boats make up a case and 4 cases make up a box. If a toy shop comes to pick up what is available on the fourth day and finds an extra 8 boxes left for them that were forgotten from a previous pickup, how many boxes of toy boats will they be able to take?\rule{0pt}{2.5ex} & 40 \\
\hline
47 & \rule{0pt}{2.5ex}The highest temperature ever recorded on Earth was 136 degrees Fahrenheit and the coldest temperature ever measured was -126 degrees Fahrenheit. If the average temperature of Earth is 59, what would be the difference between the average temperature on Earth and the average given the two extremes?\rule{0pt}{2.5ex} & 54 \\
\hline
48 & \rule{0pt}{2.5ex}Maria was shopping for the perfect prom dress. She found a red one that cost \$250 but was on sale for 20\% off. Sales tax is 5\%. Her grandmother gave her \$300 to pay for her dress and dinner. After Maria purchased the red dress, how much did she have left, in dollars, to pay for dinner?\rule{0pt}{2.5ex} & 90 \\
\hline
49 & \rule{0pt}{2.5ex} Mrs. Watson, a high school Spanish teacher, is required to input 2 grades a week per student per her school's grading policy. Mrs. Watson has 6 classes in total. Her 1st-period class has 32 students, 2nd-period has 28 students, 3rd-period has 41 students, 4th-period has 23 students, 5th-period has 18 students, and her 6th-period class has 33 students. How many grades does Mrs. Watson need to input each week to remain compliant with the school's grading policy?\rule{0pt}{2.5ex} & 350 \\
\hline
50 & \rule{0pt}{2.5ex}Mary sells 4 bags of pears where each bag contains 3 giant pears and 4 small pears. Each giant pear is sold at \$5 and each small pear is sold at \$2. Mary also sells 2 bags of cherries where each bag contains 5 pounds of cherries. The cherries are sold at \$8 per pound. How much in total does Mary earn from selling all these fruits?\rule{0pt}{2.5ex} & 172 \\
\hline
\end{longtable}

%% file: paper/xy_plots.tex
\begin{figure}[ht!]
  \includegraphics[width=\linewidth]{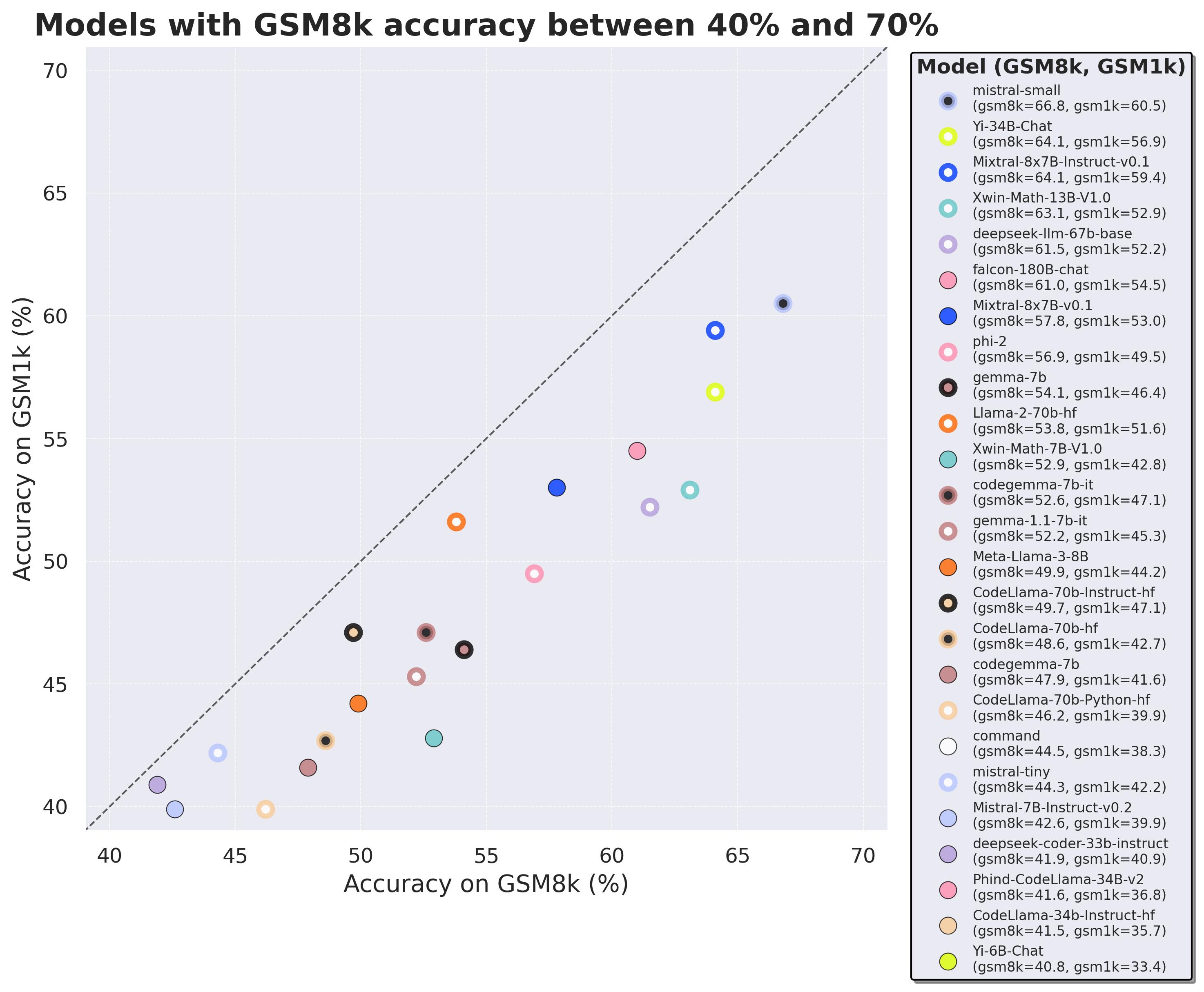}
  \caption{Models with between 40 and 70\% accuracy on GSM8k compared to the line of no overfit. This plot is zoomed into the relevant sections (40-70\% accuracy). We observe that no models lie on the line of no overfit in this regime.}
  
  \label{fig:medaccuracyplot}
  
\end{figure}

\begin{figure}[ht!]

  \includegraphics[width=\linewidth]{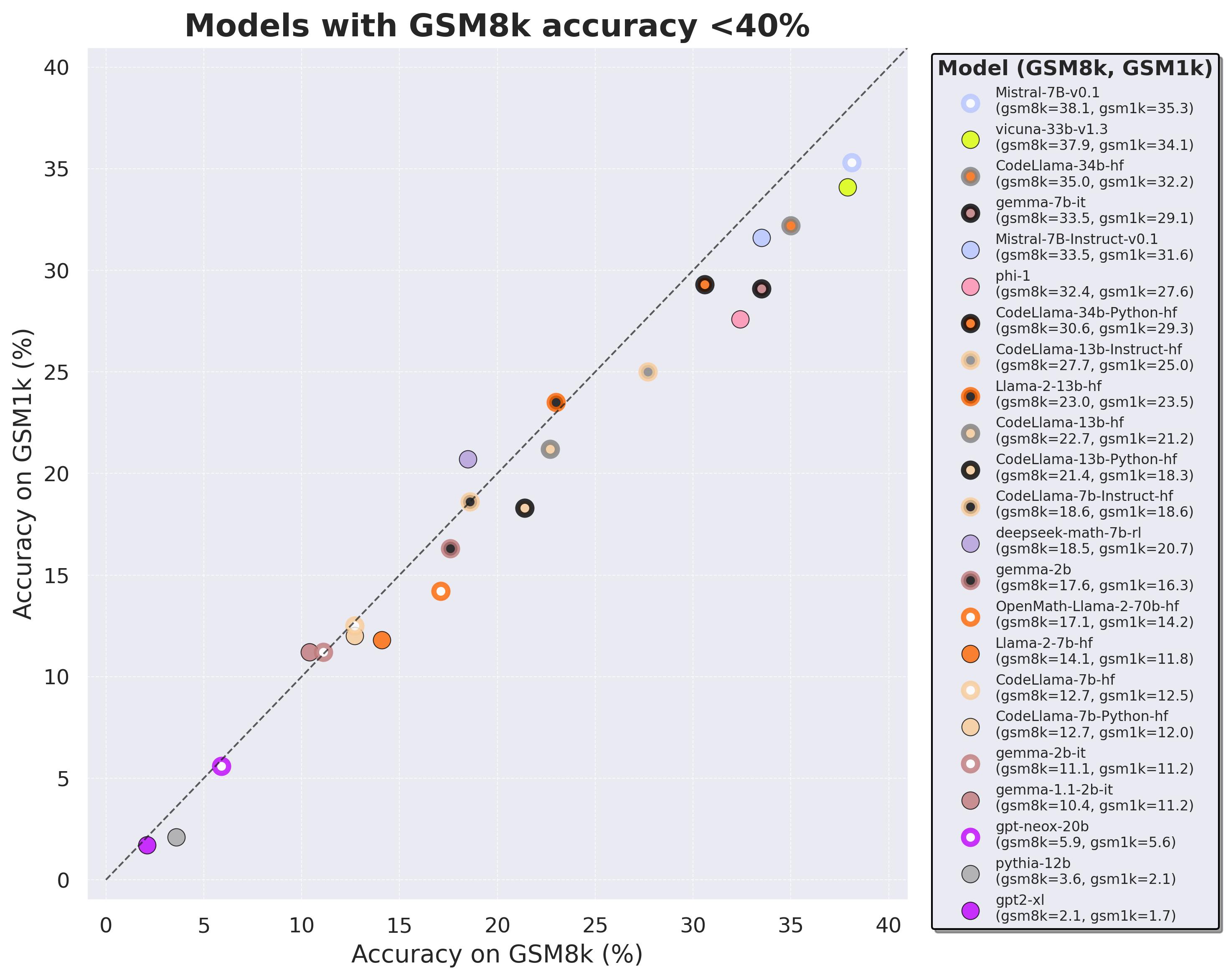}
  \caption{Models with between 0 and 40\% accuracy on GSM8k compared to the line of no overfit. This plot is zoomed into the relevant sections (0-40\% accuracy).}
  \label{fig:badaccuracyplot}
\end{figure}

%% file: paper/human_extraction.tex
\section{Ablation with Human Answer Extraction}
\label{app:manual}

Because LM Evaluation Harness uses an automatic extraction format which takes the last number outputted as the model's final answer and compares an extract string match with the gold standard. Models answers which fail to follow the proper format may be marked as incorrect even if the model produced the ``correct'' answer. This most frequently occurs when the model phrases its answer in natural language, but adds extraneous information at the end of its sentence. Additionally, automatic extraction is brittle: in a few cases we observe that model outputs such as ``24.0'' are marked as not matching the gold standard answer of 24 due to the trailing decimal points failing the exact string match.

To measure the impact of extraction errors, we analyze a subset of models and use human annotators to extract model answers. These models were chosen for high performance on GSM1k for the purposes of creating a leaderboard of top performing models. While the absolute performance numbers change, we do not find meaningful differences in the amount of overfitting between GSM1k and GSM8k based on whether human or automatic extraction was used for this set of models.

\begin{table}[ht]
\centering
\begin{tabular}{l*{6}{c}}
\toprule
\textbf{Model} & \makecell[c]{\textbf{GSM1k} \\ \textbf{(Human)}} & \makecell[c]{\textbf{GSM1k} \\ \textbf{(Auto)}} & \makecell[c]{\textbf{GSM8k} \\ \textbf{(Human)}} & \makecell[c]{\textbf{GSM8k} \\ \textbf{(Auto)}} & \makecell[c]{\textbf{Gap} \\ \textbf{(Human)}} & \makecell[c]{\textbf{Gap} \\ \textbf{(Auto)}} \\
\midrule
Claude 3 Opus & 0.952 & 0.825 & 0.955 & 0.802 & -0.003 & 0.023 \\
GPT-4 Turbo Preview & 0.951 & 0.898 & 0.952 & 0.898 & -0.001 & 0.0 \\
GPT-4o & 0.949 & 0.928 & 0.962 & 0.933 & -0.014 & -0.005 \\
Claude 3 Sonnet & 0.933 & 0.744 & 0.926 & 0.719 & 0.006 & 0.024 \\
Gemini 1.5 Pro (post-I/O) & 0.923 & 0.895 & 0.933 & 0.915 & -0.01 & -0.02 \\
Gemini 1.5 Pro (pre-I/O) & 0.905 & 0.885 & 0.904 & 0.897 & 0.002 & -0.011 \\
Llama 3 70B Instruct & 0.901 & 0.895 & 0.917 & 0.896 & -0.016 & -0.001 \\
Gemini 1.5 Flash & 0.901 & 0.832 & 0.896 & 0.804 & 0.005 & 0.027 \\
Mistral Large & 0.875 & 0.853 & 0.892 & 0.853 & -0.017 & 0.0 \\
Gemini 1.0 Pro & 0.798 & 0.789 & 0.805 & 0.792 & -0.007 & -0.002 \\
CodeLlama 34B Instruct & 0.375 & 0.366 & 0.422 & 0.415 & -0.047 & -0.049 \\
\bottomrule
\end{tabular}
\label{tab:performance_metrics}
\end{table}

%% file: paper/katie_ablations.tex
\section{Ablations with the Alternative Format}
\label{app:ablation_alt}
We investigate the impact of prompting on several of the model families with highest amounts of overfitting. In this section, we test whether the difference in performance with the standard and alternative prompt is due to the alternative prompt using non-GSM8k examples. We do this by constructing prompts in the same format as the alternative ``chain-of-thought'' prompt but using fewshot example problems randomly chosen from GSM8k. These prompts imitate the alternative prompt's answer format and use of 8 fewshot examples rather than 5 in the standard prompt. We do this by converting two sets of randomly selected GSM8k problems into the analagous format.

We find significant variance in the results, ablating even something as simple as \emph{which} n-shot examples are chosen. Nevertheless, the general shape of the findings remains largely consistent, even if the precise ordering / numerical values are highly prompt dependent.

\begin{figure}[ht!]
    \centering
    \includegraphics[width=1\linewidth]{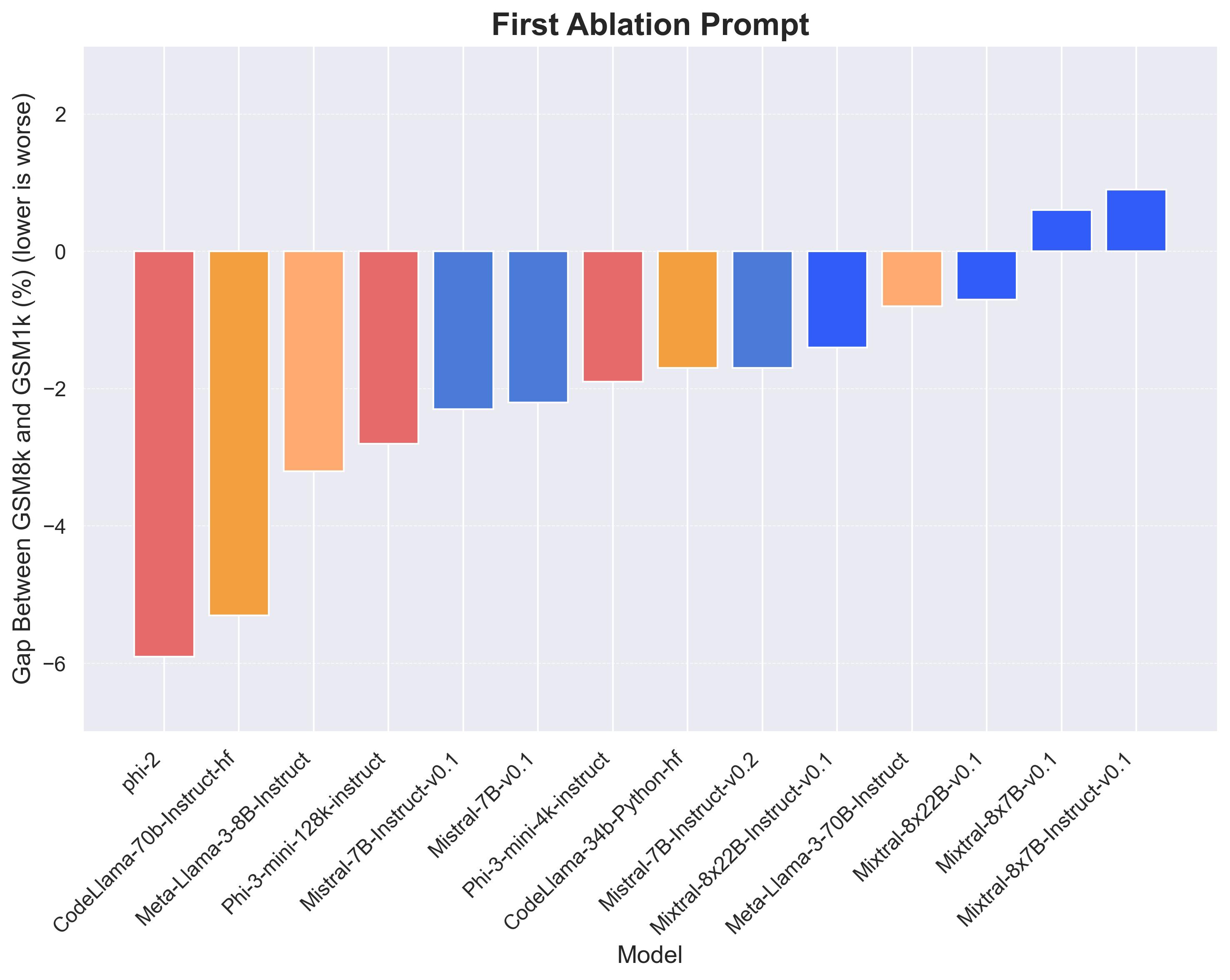}
    \caption{Models from the most overfit families arranged by their drop in performance between GSM8k and GSM1k (lower is worse) on the first ablation prompt with GSM8k examples in the alternative format.}
    \label{fig:ablation-1}
\end{figure}

For the first such ablation prompt (provided in full below), our results are displayed in Figure~\ref{fig:ablation-1}
\begin{lstlisting}
Q: Bob drove for one and a half hours at 60/mph.  He then hit construction and drove for 2 hours at 45/mph. How many miles did Bob travel in those 3 and a half hours?
A: Bob drove for 1.5 x 60 = 90 miles first, then another 2 x 45 = 90 miles. In total Bob drove 90 + 90 = 180 miles. The answer is 180.

Q: Mary is paying her monthly garbage bill for a month with exactly four weeks. The garbage company charges Mary $10 per trash bin and $5 per recycling bin every week, and Mary has 2 trash bins and 1 recycling bin. They're giving her an 18%
A: Every week, Mary pays 10 x 2 = 20 dollars for the trash bins, so her total weekly cost is 20 + 5 = 25 dollars for the trash bins and the recycling bin. Then her monthly cost over 4 weeks is 25 x 4 = 100 dollars. Mary's senior discount is 18 x .01 x 100 = 18 dollars. So subtracting the discount and adding the fine, her total monthly cost is 100 - 18 + 20 = 102 dollars. The answer is 102.

Q: June has $500 for buying school supplies for the new school year. She buys four maths books at $20 each, six more science books than maths books at $10 each, and twice as many art books as maths books at $20 each. If she also bought music books, how much money did she spend on music books?
A: The total cost of maths books is 4 x 20 = 80. She bought six more science books than maths books which totals 6 + 4 = 10 books. If each science book cost her $10 she spent 10 x 10 = 100 dollars on science books. There were twice as many art books as maths books which total 2 x 4 = 8. The total cost for art books is 8 x 20 = 160. The total amount that she used for maths, science, and art books is 160 + 100 + 80 = 340. The amount she spent on music books is 500 - 340 = 160. The answer is 160.

Q: A play was held in an auditorium and its ticket costs $10. An auditorium has 20 rows and each row has 10 seats. If only 3/4 of the seats were sold, how much was earned from the play?
A: There are 20 x 10 = 200 seats in the auditorium. Only 200 x 3/4 = 150 seats were sold. Hence, the earnings from the play is 10 x 150 = 1500. The answer is 1500.

Q: Brendan went fishing with his dad. Brenden caught 8 fish in the morning. He threw 3 back that were too small. He caught 5 more in the afternoon. Brendan's dad caught 13 fish. How many fish did they catch in all?
A: Brenden caught 8 fish in the morning and 5 in the afternoon so he caught 8 + 5 = 13 total fish. After throwing the small fish back, Brenden has 13 - 3 = 10 fish. Together, Brenden and his dad caught 10 + 13 = 23 fish. The answer is 23.

Q: Valerie's cookie recipe makes 16 dozen cookies and calls for 4 pounds of butter.  She only wants to make 4 dozen cookies for the weekend.  How many pounds of butter will she need?
A: Her original recipe makes 16 dozen and she only needs 4 dozen so she needs to reduce the recipe by 16 / 4 = 4. For 4 dozen cookies, she needs to reduce her recipe by 4 and the original called for 4 pounds of butter so she now needs 4 / 4 = 1 pound of butter. The answer is 1.

Q: Jack is mad at his neighbors for blasting Taylor Swift all night, so he slashes three of their tires and smashes their front window. If the tires cost $250 each and the window costs $700, how much will Jack have to pay for the damages?
A: The total cost of the tires is 250 x 3 = 750. Then total cost of the tires and the window is 700 + 750 = 1450. The answer is 1450.

Q: A dental office gives away 2 toothbrushes to every patient who visits.  His 8 hour days are packed and each visit takes .5 hours.  How many toothbrushes does he give in a 5 day work week?
A: Each day he does 8 / .5 = 16 visits. So he does 16 x 5 = 80 visits a week. That means he gives away 80 x 2 = 160 toothbrushes a week. The answer is 160.
\end{lstlisting}

\begin{figure}[ht!]
    \centering
    \includegraphics[width=1\linewidth]{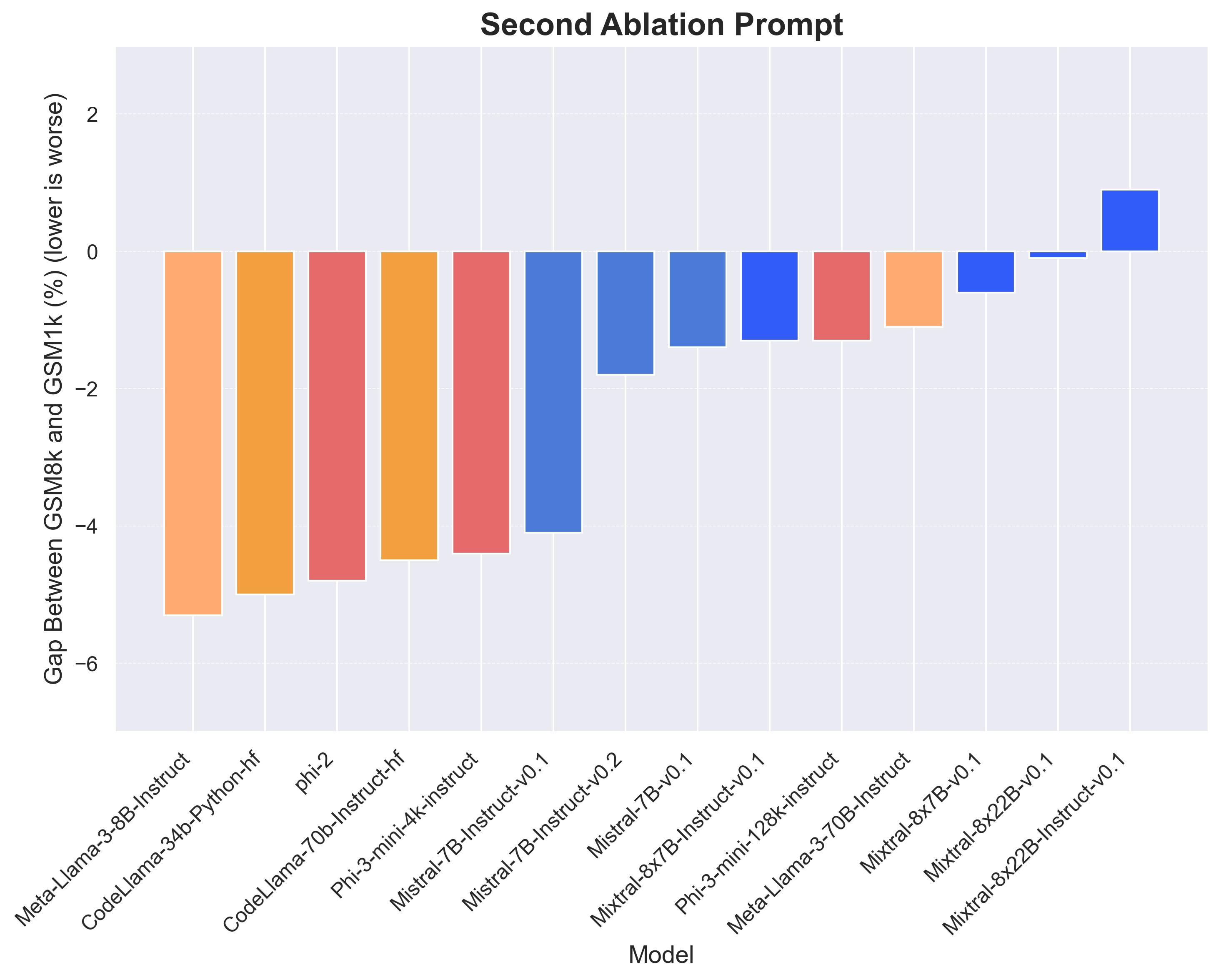}
    \caption{Models from the most overfit families arranged by their drop in performance between GSM8k and GSM1k (lower is worse) on the second ablation prompt with GSM8k examples in the alternative format.}
    \label{fig:ablation-2}
\end{figure}

For the second such ablation prompt (provided in full below), our results are displayed in Figure~\ref{fig:ablation-2}
\begin{lstlisting}
Q: At camp Wonka, there are 96 campers.  Two-thirds of the campers are boys, and the remaining one-third are girls.  50%
A: The girls make up one-third of the campers, so there are 96 / 3 = 32 girls. The boys make up two-thirds of the campers, so there are 32 + 32 = 64 boys. There are 32 x 75%

Q: In today's field day challenge, the 4th graders were competing against the 5th graders.  Each grade had 2 different classes.  The first 4th grade class had 12 girls and 13 boys.  The second 4th grade class had 15 girls and 11 boys.  The first 5th grade class had 9 girls and 13 boys while the second 5th grade class had 10 girls and 11 boys.  In total, how many more boys were competing than girls?
A: When you add up all the girls from all 4 classes, you had 12 + 15 + 9 + 10 = 46 girls. When you add up all the boys from all 4 classes, you had 13 + 11 + 13 + 11 = 48 boys. There are 48 boys and 36 girls so 48 - 46 = 2 more boys. The answer is 2.

Q: Axel bought an aquarium that was marked down 50%
A: The aquarium was bought for $120 x 50/100 = $60 less. So the marked down price of the aquarium was $120 - $60 = $60. Axel paid $60 x 5/100 = $3 additional for the sales tax. Therefore, the total cost of the aquarium was $60 + $3 = $63. The answer is 63.

Q: There are 48 crayons in the box.  Kiley takes 1/4 of them away.  Joe takes away half of the remaining crayons, how many crayons are left?
A: Kiley takes 48 / 4 = 12 crayons, so 48 - 12 = 36 crayons remain. Joe takes 36 / 2 = 18 crayons, so there are 36 - 18 = 18 crayons left. The answer is 18.

Q: Six Grade 4 sections launched a recycling drive where they collect old newspapers to recycle. Each section collected 280 kilos in two weeks. After the third week, they found that they need 320 kilos more to reach their target. How many kilos of the newspaper is their target?
A: In a week, each section collected 280 / 2 = 140 kilos of newspapers. So, in three weeks, one section collected 140 x 3 = 420 kilos. So, the four sections collected a total of 420 x 4 = 1680 kilos. Hence, their target is to collect 1680 + 320 = 2000 kilos of the newspaper. The answer is 2000.

Q: Jeff has a shelter where he currently takes care of 20 cats. On Monday he found 2 kittens in a box and took them to the shelter. On Tuesday he found 1 more cat with a leg injury. On Wednesday 3 people adopted 2 cats each. How many cats does Jeff currently have in his shelter?
A: Counting the cats he had, the kittens he found, and the injured cat, Jeff had a total of 20 + 2 + 1 = 23 cats. 3 people took a total of 3 x 2 = 6 cats. After Wednesday, Jeff was left with 23 - 6 = 17 cats. The answer is 17.

Q: Paul is working at a university. He is part of a big project, which involves 70 scientists in total. Half of them are from Europe and one-fifth are from Canada. The rest are from the USA. How many scientists in this project are from the USA?
A: Of all the scientists taking part in the project, half of them are from Europe, which means 70 x 0.5 = 35 people. The number of researchers from Canada is 70 x 1/5 = 14 people. That means there are 70 - 35 - 14 = 21 researchers from the USA. The answer is 21.

Q: John buys a heating pad for $30.  He uses it 3 times a week for 2 weeks.  How much does he spend on each use?
A: He uses it 3 x 2 = 6 times. So he pays $30 / 6 = $5 for each use. The answer is 5.
\end{lstlisting}

\clearpage

\section{Ablations with Number of Fewshot Examples}
\label{app:ablation_n}
\begin{figure}[ht!]
    \centering
    \includegraphics[width=1\linewidth]{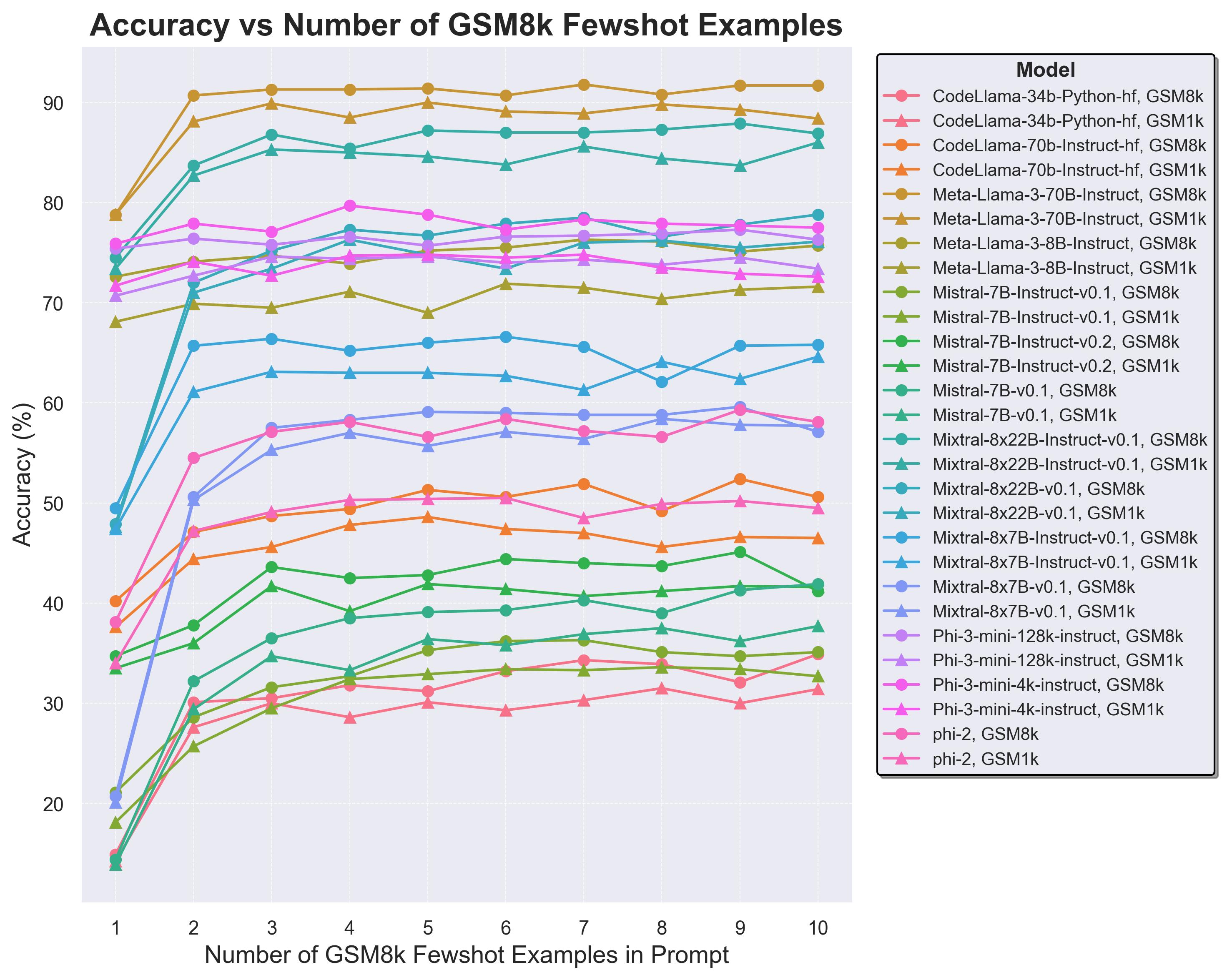}
    \caption{Performance on GSM8k and GSM1k relative to the number of GSM8k fewshot examples given in the standard prompt format, for models from the most overfit model families.}
    \label{fig:num-fewshot}
\end{figure}

As another ablation, we also investigate the impact of varying the number of GSM8k fewshot examples in the standard prompt format. We evaluate the models from the most overfit model families on the standard n-shot prompt, with n varying from 1 to 10, inclusive. As in the primary results, the n fewshot examples in the prompt are randomly selected from GSM8k train, and vary among the questions from the GSM1k/GSM8k test set. Note that the primary findings correspond to n=5. Our results are displayed in Figure~\ref{fig:num-fewshot}. We notice that all models display a performance gap between GSM8k and GSM1k for almost all values of n, demonstrating a robustness to the overfitting result despite the variance from using different prompts.

